%% file: main.tex
\documentclass[sn-mathphys,Numbered]{sn-jnl}

\usepackage{graphicx}%
\usepackage{multirow}%
\usepackage{amsmath,amssymb,amsfonts}%
\usepackage{amsthm}%
\usepackage{mathrsfs}%
\usepackage[title]{appendix}%
\usepackage{xcolor}%
\usepackage{textcomp}%
\usepackage{manyfoot}%
\usepackage{booktabs}%
\usepackage{algorithm}%
\usepackage{algorithmicx}%
\usepackage{algpseudocode}%
\usepackage{listings}%
\usepackage{float}
\usepackage[switch]{lineno}

\usepackage [latin1,utf8]{inputenc}
\usepackage{listingsutf8}

\usepackage{lscape}
\usepackage{xltabular}
\usepackage{color}
\usepackage{longtable}
\usepackage{comment}

\usepackage{thmtools}
\declaretheoremstyle[bodyfont=\itshape]{normalbody}
\declaretheorem[style=normalbody,name=Example]{example}

\definecolor{mygreen}{rgb}{0,0.6,0}
\definecolor{mygray}{rgb}{0.5,0.5,0.5}
\definecolor{mymauve}{rgb}{0.58,0,0.82}

\begin{document}

\title[Legal Summarisation through LLMs]{Legal Summarisation through LLMs: The PRODIGIT Project}

\author[1,5]{\fnm{Thiago} \sur{Dal Pont}}\email{thiagordalpont@gmail.com}
\equalcont{These authors contributed equally to this work.}

\author[1,2]{\fnm{Federico} \sur{Galli}}\email{federico.galli7@unibo.it}
\equalcont{These authors contributed equally to this work.}

\author[4]{\fnm{Andrea} \sur{Loreggia}}\email{andrea.loreggia@unibs.it}
\equalcont{These authors contributed equally to this work.}

\author*[1,2]{\fnm{Giuseppe} \sur{Pisano}}\email{g.pisano@unibo.it}
\equalcont{These authors contributed equally to this work.}

\author[1,3]{\fnm{Riccardo} \sur{Rovatti}}\email{riccardo.rovatti@unibo.it}
\equalcont{These authors contributed equally to this work.}

\author[1,2]{\fnm{Giovanni} \sur{Sartor}}\email{giovanni.sartor@unibo.it}
\equalcont{These authors contributed equally to this work.}

\affil[1]{\orgdiv{Alma-AI}, \orgname{University of Bologna}, \orgaddress{\street{Via Galliera 3}, \city{Bologna}, \postcode{40121},  \country{Italy}}}

\affil[2]{\orgdiv{Department of Legal Studies}, \orgname{University of Bologna}, \orgaddress{\street{Via Zamboni 27-29}, \city{Bologna}, \postcode{40126 }, \country{Italy}}}

\affil[3]{\orgdiv{Department of Electrical, Electronic, and Information Engineering ``Guglielmo Marconi''}, \orgname{University of Bologna}, \orgaddress{\street{Viale Risorgimento 2}, \city{Bologna}, \postcode{40136}, \country{Italy}}}

\affil[4]{\orgdiv{Department of Information Engineering}, \orgname{University of Brescia}, \orgaddress{\street{Via Branze 38}, \city{Brescia}, \postcode{25123}, \country{Italy}}}

\affil[5]{\orgdiv{Department of Automation and Systems Engineering}, \orgname{Federal University of Santa Catarina}, \orgaddress{\street{Trindade}, \city{Florianopolis}, \postcode{88040-900}, \country{Brazil}}}

\abstract{

We  present some initial results of a large-scale Italian project called PRODIGIT which aims to support  tax judges and lawyers through digital technology, focusing on AI. We have focused on  generation of summaries of judicial decisions and on the extraction of related information, such as  the identification of legal issues and decision-making criteria, and the specification of keywords. To this end, we have deployed and evaluated different tools and approaches to extractive and abstractive summarisation. We have applied LLMs, and particularly on GPT4, which has enabled us to obtain results that  proved satisfactory, according to an evaluation by expert tax judges and lawyers. On this basis, a prototype application is being built which  will  be made publicly available.
\vspace{-0.8em}
}


\maketitle

\section{Introduction}\label{sec:intro}
\input{Sections/1_Intro}

\section{Tax Law Adjudication}\label{sec:adjudication}
\input{Sections/2_TaxLawAdjudication}

\section{The PRODIGIT Dataset}\label{sec:dataset}
\input{Sections/3_DatasetTaxLaw}

\section{Summarisation of Tax Law Decisions}\label{sec:legal-summarization}
\input{Sections/4_Summarisation}

\section{NLP Summarisation Tools}\label{sec:generativetools}
\input{Sections/5_UsedTools}

\section{Extractive Summarisation}\label{sec:extractive-summarisation}
\input{Sections/6_ExtractiveSummarisation}

\section{Abstractive Summarisation}\label{sec:abstractive-summarisation}
\input{Sections/7_AbstractiveSummarisation}

\section{Evaluation}\label{sec:evaluation}
\input{Sections/8_Evaluation}

\section{Related Work}\label{sec:related-work}
\input{Sections/9_RelatedWorks}


\section{Conclusion}\label{sec:conclusion}
\input{Sections/10_Conclusions}

\bibliography{main}

\backmatter

\section*{Acknowledgments}
\input{Sections/11_Acknowledgments}


\appendix

\section{Appendix: Prodigit Summaries}\label{app:original}
\input{Sections/Appendix}

\end{document}

%% file: Sections/1_Intro.tex
The law is typically a natural-language-based domain, and natural-language texts are pervasive in the law. First,  natural language is the medium that legislation (including administrative regulations of all kinds) uses to express legal prescriptions, which humans (both experts and laypeople) are assumed to understand and comply with. Legislative and regulatory bodies have produced complex and evolving networks of natural language texts, which have complex structures and interconnections and use diverse terminologies to express technical and non-technical content. Second, natural language is used in judicial proceedings and opinions. In a proceeding, the parties to a legal case rely on natural language to express their arguments, motions, and claims, as do witnesses in their testimonies.  In their opinions, judges use natural language to report the facts of the case, summarise the arguments made by the parties, and express the reasons behind interpretations, rulings, and decisions. Natural language is normally used by private parties to express contracts and other agreements as well as the accompanying documents. Finally, natural language is the usual medium for a host of legally relevant documents, at any level of complexity, that are relevant for the creation, interpretation, and application of the law, such as doctrinal scholarship, legal theories, and case commentaries. 

Natural language in a legal text may use complex syntactical structures and rich terminologies, whose dense meaning results from the combination of common sense, technical knowledge, and past legal interpretations. The complexity and density of legal language have so far been a key obstacle to the deployment of AI technologies. This has been the case, on the one hand, for the deployment of symbolic AI approaches, which have struggled to capture, through the chosen formalisms, the variety, ambiguity, and meaning density of legal language. The more such formalisms have tried to reproduce the richness of natural language, the more work-intensive the knowledge-representation exercise has been and the more debatable its outcomes. 
On the other hand, the complexity of legal language has also limited the  application of machine-learning NLP-based approaches. Most commonly, a supervised approach has been adopted where the links between input texts and the targets being predicted have to be specified by manually tagging large training sets of legal documents. The preparation of such training sets is very labour-intensive, and the information to be extracted is limited to the connections emerging from the tags in the training set.

This scenario may now be changed by the advent of large language models, also called ``foundational models''\cite{BommasaniHudsonOthers2022OR}. The largest and best performing family of such models so far is the GPT family, developed by Open AI, the latest releases of which are GPT3.5 and GPT4 (both embedded in ChatGPT) \cite{openai2023gpt4}. Other large language models have recently been produced, an example being Google's Bard. As is known, these models are very large neural networks, operating on hundreds of billions of parameters (links in the network). They have been constructed by training such networks on enormous sets of natural language documents. Based on that, a network learns to predict the texts (sequences of words) that are likely to meaningfully complement the input contexts (the prompts) that are  provided by users (for an introduction to LLMs, see \cite{Wolfram2023WI}). 

The results achievable through large language models have been surprisingly good. The ``stochastic parrots" -- as these systems have been called given their ability to express well-formed and seemingly meaningful language without having a real understanding of what is being said \cite{BenderGebruMcMillan-MajorShmitchell2021On} -- can draft complex texts that in many cases have sufficiently good quality for different uses relevant to the law (translation, summarisation, document analysis, draft generation, completion).
It is true that in some cases such systems ``hallucinate'', i.e., they provide content (e.g., legal citations or reasoning) that does not match reality, or give answers that violate basic logic. However, even if they only mimic the text generation of cognisant humans, their performance is considerable and in many cases surprisingly good, or at any rate satisfactory for many practical applications.

In the following, we shall discuss how LLMs have been used in the PRODIGIT project, an initiative developed in Italy by the Presidential Council of Tax Justice (Consiglio di Presidenza della Giustizia Tributaria, CPGT), in cooperation with the Ministry of the Economy and  Finance (Ministero dell'Economia e delle Finanze, MEF), and funded by the National Operative Project for Governance and Institutional Capacity Programme 2014-2020. The general aim of the project is to provide support to tax judges and lawyers through digital technologies, in particular through AI techniques.
In the project, LLMs have been used for two main purposes: (1) to prepare summaries and headnotes of judicial decisions and extract related information; and
(2) to provide semantic tools for searching and analysing the case law.

In the following, we shall first describe the context of the project (Section~\ref{sec:adjudication}) and the dataset used (Section~\ref{sec:dataset}) and will then address the summary-generation function (Sections~\ref{sec:legal-summarization} to~\ref{sec:abstractive-summarisation}). 
Following this analysis, we present the evaluation procedure based on surveys carried out with legal experts (Section~\ref{sec:evaluation}). Finally, we consider related work on the use of LLMs in the legal domain (Section~\ref{sec:related-work}) and provide some considerations for future work (Section \ref{sec:conclusion}).

%% file: Sections/2_TaxLawAdjudication.tex



Italian tax law adjudication involves three levels: (1) tax courts of first instance, (2) tax courts of second Instance, and (3) the Supreme Court of Cassation. 

The process begins with the tax courts of first instance, where taxpayers can file complaints against the decisions made by central and local Italian tax authorities. These courts have jurisdiction over cases related to most tax matters, including income tax, value-added tax (VAT), corporate tax, and local taxes. Both parties (taxpayers and tax authorities) present their arguments and evidence during the first instance proceedings, after which  the court issues a legally binding judgement. If dissatisfied, either the taxpayer or the tax authority or tax office can appeal to a tax court of second instance, which have jurisdiction over an entire region. These have the power to confirm, modify, or overturn the first-instance decision. Further appeals can be made to the Supreme Court of Cassation, the highest judicial authority, on matters of law (the Court of Cassation cannot re-examine the facts of the case). Tax law judges are a mixed group, including both professional  and non-professional judges. The latter are usually lawyers or accountants who  serve part-time in a judicial capacity and are paid based on the number of cases they decide. The quality of their decisions is often said to vary significantly, and to be on average lower than in other parts of the judiciary.

Italy, like other countries, faces a very large number of tax-related cases. This is determined by many factors, among which the large taxpayer base, the relatively low level of tax compliance, and the uncertainties in tax law, as determined by the complexity of the tax regime and the frequent changes in tax law provisions.
The 2022 report provided by the Department of Finance\footnote{MEF Dipartimento delle Finanze, Relazione sul Contenzioso della Giustizia Tributaria, giugno 2023 \url{https://www.finanze.gov.it/export/sites/finanze/.galleries/Documenti/Contenzioso/Relazione-monitoraggio-contenzioso-2022.pdf}} shows that the number of complaints received by tax courts of first instance in 2022 was about 145,972, while appeals brought to courts of second instance were 41,051. The Court of Cassation receives about 10,531 appeals against second-instance decisions.

Italy has made some efforts to modernise its tax adjudication system by embracing digital solutions. One significant development is the introduction of electronic filing and communication systems, referred to as a ``telematic tax process'' (\textit{Processo Tributario Telematico}), which has allowed for the almost complete digitisation of all stages of the judicial process. Through this system, taxpayers can submit appeals, supporting documents, and relevant information, and judges can read, write, and deliver their measures electronically, eliminating the need for physical paperwork, time-consuming procedures, and administrative burdens. In particular, with regard to the decision-making phase, the platform for preparing a  judicial decision in digital form (referred to as a ``digital judicial decision'' -- \textit{Provvedimento Giurisdizionale Digitale}, or PGD) has been fully operational in all tax courts since 1 December 2021 and allows the judge to draft, sign, and file decisions fully electronically. At present, the PGD enables the electronic drafting of collegiate (panel) judgements, as well as orders and judgements issued by a sole judge.

New technologies, including machine learning and LLMs, are starting to be used to address certain aspects of tax administration. For instance, data analytics and AI tools are currently employed by the Italian Tax Administration to analyse large bodies of financial and transactional data and detect potential discrepancies or irregularities.\footnote{Bloomberg Tax, Italy Turns to AI to Find Taxes in Cash-First, Evasive Culture, available at \url{https://news.bloombergtax.com/daily-tax-report-international/italy-turns-to-ai-to-find-taxes-in-cash-first-evasive-culture}} Hopefully, this will allow for a more efficient and targeted approach to tax audits and investigations.
Tax adjudication is lagging behind in the use of AI technologies by comparison with tax administration.
The PRODIGIT project aims to make AI technologies available in tax adjudication as well, so as to provide judges and professionals with better, more targeted information and help them efficiently address the complexities of tax law. 

%% file: Sections/3_DatasetTaxLaw.tex
The PRODIGIT project aims to provide tools that can be applied to the whole of Italian case law in the tax domain.
However, for the purpose of experimentation and prototyping, a restricted domain was selected, namely, decisions concerning the registration and recordation tax (\textit{imposta di registro}). This tax concerns the registration and recordation of deeds and other legally relevant documents and applies in particular to various kinds of contracts (such as those involving the transfer of real estate). It has the dual purpose of providing tax revenue and of paying the state for the service it provides to private individuals, namely, keeping track of particular deeds and financial transactions to give them legal certainty. It is governed by Presidential Decree No 31/1986 (\textit{Testo Unico dell'imposta di registro}).

We only considered those decisions that have been produced in a native digital format through the online platform provided to tax judges. In the future, the data set will be expanded to also include the vast amount of past decisions that are currently only available as scanned images of paper documents.

We started with a collection of approximately 1,500 decisions addressing certain selected topics within the domain of the registration tax. 
Of such decisions, 750 were delivered by tax courts of first instance and 712 by tax courts of second instance. These decisions span between 2021 and 2023, with most decisions issued in 2022. They have been delivered by the tax courts of different Italian regions and provinces.

The decisions have a standard structure consisting of the following parts:

\begin{enumerate}
 \item \textit{Introduction}, reporting (i) the number of the decision, (ii) the composition of the judicial panel, (iii) the parties and their attorneys (if present), the latter of which had previously been anonymised; 
 \item \textit{Development of the Proceeding}, reporting (a) the facts related to the tax administrative process and, when delivered at the second instance (on appeal), the procedural facts related to the first-instance (trial court) proceedings (e.g., the parties' requests, claims, and arguments, as well as first-instance decisions by the tax court); (b) the requests by the parties, often presented with the related claims and arguments, possibly formulated as appeals against the first-instance decision;
\item \textit{Grounds of the Decision}, stating the reasons in fact and in law supporting the court's decision;
 \item \textit{Final Ruling}, stating whether the complaint or appeal has been accepted or denied, and allocating the costs of the proceedings.
\end{enumerate}

A dataset of 17,000 decisions from various areas of tax law is currently being normalised -- i.e., anonymised, segmented into relevant partitions, and corrected to fix typos and garbled text -- and will be used in future developments of the project.

%% file: Sections/4_Summarisation.tex
The first task addressed in the PRODIGIT project concerns the summarisation of judicial decisions. In this section, we introduce the concept of summarisation and present the running example that will be used in the following section
\subsection{Summarisation in the Legal Domain }

Summarisation is the process of condensing a large set of input information into a shorter document, the summary, which still contains the most significant information, or at any rate the information that is relevant to the task at hand. In general, summarisation is subject to the need to jointly satisfy conflicting requirements as best as possible: providing a summary that is as short as possible but still  includes as much of the relevant information as possible.

Summarisation is very important in the legal domain, where the amount of available legal materials overwhelms the human capacity to process them. By providing summaries of decisions, judges and lawyers are given a chance to determine more quickly whether a precedent is relevant to the issue at hand, and decide whether it is worth their while engaging with the text in its entirety. Moreover, summarisation may highlight the key points of a lengthy decision, enabling lawyers to focus on them. Legal documents are particularly challenging for summarisation compared to other types of texts. These challenges relate to multiple aspects, such as the length of the documents, the hierarchical and interconnected structure of their parts, theit complex technical vocabulary, and the ambiguity of natural legal language, as well as the importance of citations to legal sources.

In Italian legal culture, we can distinguish two kinds of summary accounts (or statements) of judicial decisions.

The first account consists of so-called ``maxims'' (\textit{massime} in Italian). A maxim (\textit{massima}) specifies the most significant principles stated in leading judicial decisions. There exists an office in the Italian Supreme Court (the \textit{Ufficio Massimario}) that is tasked with preparing maxims from the case law of that court. The highly qualified judges working in this office identify what decisions deserve a maxim, since they introduce  principles that are particularly important, establishing new law or solving a previously unsettled issue. These principles are given compact linguistic formulations (the maxims) which are published in an online collection.  In the tax domain, the function of preparing maxims has until recently been carried out by regional bodies and will be entrusted to a national body in the future. Maxims are important in the Italian legal system since they are often taken as authoritative statements of the law, and are used in arguments to support interpretive and other claims. There is indeed a debate among Italian lawyers on the extent to which maxims effectively contribute to a knowledge of the law and to legal certainty, capturing with precision the underlying rationale or \textit{ratio decidendi} of important cases. In any event, this is an important and persisting aspect of Italian legal culture.

The second account consists, more modestly, in providing summaries, i.e., abstracts of legal cases, to be used to ``triage'' retrieved cases and identify the points in them that are most relevant. In other words, a summary enables lawyers decide whether they should engage with the whole case (or a section of it) and points them to its most significant aspects.  It is this second kind of account (the summary) that we aim to provide with PRODIGIT. We do not undertake to replace the production of maxims, a task that, as noted, requires advanced legal skills, with the maxims themselves playing a distinctive role and institutional arrangement in the Italian legal system.  However, we believe that automated summarisation, particularly in the form of the automated extraction of ``decision-making criteria'' (see Section \ref{sec:abstractive-summarisation} may provide useful support to the office tasked with preparing the maxims, the \textit{Ufficio Massimario}).

We experimented with both extractive and abstractive summarisation, considering that both approaches are potentially useful in the legal domain, and indeed they present complementary advantages and disadvantages.

Extractive summarization selects the most meaningful sentences in the input text and combines them to form the summary. No change is made to the textual content of the extracted sentences. The extractive approach has the advantage of ensuring that all content in the summary is obtained from the input document, without any spurious addition. Moreover, it enables the reader to move from the selected sentences to their position in the original document, so as to obtain a context for such sentences, when needed. On the other hand, the extractive approach may fail to capture all relevant content or may do so only at the cost of reproducing large parts of the original texts, thus defeating the very purpose of summarization.

Abstractive summarization generates a new text which aims to provide a synoptic statement of the content of the input documents, without reproducing their wording. The abstractive approach -- when it does its job well -- has the advantage of providing a short text that, in an appropriate linguistic form, still captures the salient content of a much larger document. But it may not work well, so it carries the risk of misleading readers by ``hallucinating'', generating content that is not found in the original text.

Both extractive and abstractive summarization can find their use in the legal domain. 
The extractive approach is most appropriate in those cases in which the input documents are well-structured, being  divided into relatively short sentences, each of which delivers a separate message. This is usually the case with decisions by the highest courts, which generally address separately the different issues submitted by the parties (the claims against lower-level decisions being contested), presenting in an orderly manner the reasons for deciding the case in one way or the other. These high courts see it as part of their mission to state binding or at least persuasive principles that should guide lower courts, and they often expressly state these principles in separate, well-recognisable sentences. Unfortunately, this was not the case with the decisions we considered, namely, first-instance and appeal decisions in the tax domain, which often include long sentences addressing different issues, often introducing new elements that concern issues discussed in previous sentences.

The abstractive approach may be more appropriate when the legal reasoning is developed in long, sprawling sentences, having mixed content, or when a long decision needs to be summarised into a short and clear  account,  regardless of the way in which this content is expressed in the original document. In our experiments, we used both extractive and abstractive approaches and submitted their outcome to the evaluation of legal experts (see Section \ref{sec:evaluation}).

\subsection{Running Example}\label{rexample}

As a running example, we use Decision No. 7683 issued on 14 September 2022 by the Court of Second Instance of Sicily.

The case concerns the application of the first-time home buyer tax relief on the purchase of a house by a person who already owns another property. The buyers argued that they were entitled to the relief since that property was unsuitable for housing. 

The tax office had nevertheless refused to grant the relief, so the buyer attacked this decision in front of the tax court of first instance, winning the case, i.e., finding that the buyer was entitled to the relief.

The tax administration appealed the decision, and the court of second instance upheld the first-instance decision on the ground that the property already owned by the buyer was unsuitable for housing according to an expert home-inspection report. This was argued based on the case law of the Supreme Court, according to which the law on the registration tax has to be interpreted in such a way that the tax relief is to be denied only to those who own a house concretely suitable to be used as a dwelling.

In the following, we shall use this case both for extractive and abstractive summarisation, which we will illustrate by presenting some portions of (the English translations of) the outputs of our experiments. The entire text of the case and some summaries being produced  are available in the Appendix \ref{app:original}, in an English translation.  The original outcomes of our experiments, in Italian, as well as further outputs, are available in this  online repository.

The implementation was conducted within the IBM Cloud Pak for Data environment\footnote{https://www.ibm.com/products/cloud-pak-for-security - Accessed on August 1st, 2023}. This platform offers comprehensive packages essential for AI-based solutions. To access OpenAI models, we utilized the Microsoft Azure API\footnote{https://azure.microsoft.com/ - Accessed on August 1, 2023}.


%% file: Sections/5_UsedTools.tex

Natural Language Processing is a flourishing field with methods ranging from classical statistical approaches to very large ML-based models capturing subtle semantic features.

In our research, we tested both special-purpose NLP tools for extractive summarization and the recent large language models, which we deployed for both extractive and abstractive summarization. 

\subsection{Special-Purpose NLP Tools}\label{subsec:nlptools}

Extractive summarization techniques have been around for several decades and are widely used, including in the legal domain. As noted above, the goal of these methods is to extract the most significant sentences from a text with multiple paragraphs, under the assumption that such selected sentences can sum up the meaning of the entire text, or at least convey its most significant legal content.

As a first option, we tackled extractive summarization using the following established techniques.

\paragraph{Latent Semantic Analysis (LSA)} LSA  uses singular value decomposition to identify the underlying relationships between words in a document.  It assigns a weight to each sentence based on its semantic similarity to the entire document. Sentences with greater weights are considered more important and are included in the summary. The goal is to create summaries with wide coverage of the document's main content while avoiding redundancy~\cite{Gong2001}. 

\paragraph{Lex-Rank} Lex-Rank uses a graph-based approach to identify the most important sentences in a document. It creates a graph where each sentence is a node, and the edges between the nodes represent the similarity (cosine) between sentences. The most important sentences are those that have the highest centrality scores, which are calculated using PageRank~\cite{Brin1998}. Lex-Rank is quite insensitive to the noise in the data that may result from an imperfect topical clustering of documents~\cite{Erkan2004}.

\paragraph{TextRank} This is a general-purpose graph-based ranking algorithm for NLP. Essentially, it runs PageRank on a graph designed for summarization. It builds a graph using some set of text units as vertices. Edges are based on the measure of semantic or lexical similarity between the text unit vertices. Unlike PageRank, the edges are typically non-directed and can be weighted to reflect  degrees of similarity. 

\paragraph{Luhn} Luhn uses a statistical approach to identify the most important sentences in a document. The approach assigns a score to each sentence based on the frequency of important words in the sentence. 
One advantage of Luhn is that it is a simple and interpretable algorithm that can be easily implemented. 

\paragraph{Natural Language Toolkit (NLTK)} NLTK is a platform for building Python programs to work with human language data. Its summarizer applies a variation of the TextRank algorithm. It creates a graph where each sentence is a node, and the edges between the nodes represent the similarity between sentences. It assigns a score to each sentence based on its centrality in the graph. 
The NLTK summarizer is easy to use and can generate high-quality summaries. However, the NLTK summarizer may struggle to handle documents with complex language or a large number of irrelevant sentences.

\subsection{Large Language Models}\label{subsec:llms}

Besides exploring task-specific techniques, we also explored the applicability of more general-purpose tools like Large Language Models (LLMs), among which IT5 and GPT.

LLMs use transformer architectures based on a combination of large neural networks with attention mechanisms to track relationships between the words in a text.
Transformers are pre-trained on large amounts of text in a self-supervised fashion, i.e., they are automatically trained from the input text without human intervention. In this way, they learn the statistical connections between words. In their sequences, a piece of information is used to develop the basic ability of these models, namely, to effectively predict the next words given an initial text. Though LLMs can be seen as language-to-language machines, all internal processing is numeric. Hence LLM input stages  provide embedding functions, i.e., mappings from text chunks to high-dimensional numeric vectors that reflect the semantic  features of such text chunks. 
In more advanced LLMs, the core statistical engine may be wrapped in further layers, which provide  the system with additional capacities, such as  following instructions,  producing output in a prescribed format, and 
preventing the delivery of inappropriate output.

Though, in principle, pre-trained LLMs can be fine-tuned to specific text corpora, so far we have not engaged in any further training.  This is due to the fact that  good results could be obtained without fine-tuning the general GTP4 model, and also by the fact that we did not have a large set of tax decisions summarised by humans. Some summaries have been created in the past, but their number is limited and their styles and quality vary greatly. Moreover, the improvement obtained by fine-tuning large language models so far appears to be limited and probably will be even more limited in the future, when the capabilities of state-of-the-art tools improve. However, we plan to engage in some fine-tuning experiments in future developments of this project. We will assess whether this direction is worth taking on the basis of  outcomes of these experiments.

In the following, we give further details on the LLM we employed.

\paragraph{IT5} The IT5 model family offers a sequence-to-sequence transformer model for the Italian language. Based on the Transformer-XL architecture, this model has been pre-trained on a vast dataset of over 5 million web pages, enabling it to capture long-term dependencies in the text. The model provides more coherent and consistent text compared to earlier transformer-based models. Additionally, it has the capability to generate text from a given prompt, making it well-suited for tasks such as summarization, question-answering, and dialogue generation \cite{Sarti2022it5}. One can use IT5 for such tasks by accessing the model available at \textit{Huggingface} platform\footnote{https://huggingface.co/efederici/sentence-it5-base}, which is straightforward to use with the Python programming language. Of the available models for IT5, we used both \textit{large} and \textit{small} ones.

\paragraph{GPT}
GPT, short for Generative Pre-trained Transformer, is an auto-regressive language model that employs deep learning techniques to produce text that resembles human language. This model is based on a large-scale transformer architecture and has been trained using a vast dataset of webpages.
GPT uses a deep neural network to generate text that is highly similar to human writing. The model is trained in a self-supervised learning task, which involves predicting the next word in a sequence of words, given all the previous words. As a result, GPT can produce coherent, fluent, and practically indistinguishable text from human-written text. It can be leveraged for a wide range of natural language processing tasks, such as answering questions, summarising text, and translating languages. Those interested in using GPT can easily access the API offered by OpenAI.
We used GPT both in version 3.5 \cite{brown2020language} and version 4 \cite{openai2023gpt4}, the latter being the latest and, according to its developers, far more capable than its predecessor in tasks like sentiment analysis and text classification. GPT-4 can also process a more significant number of input and output tokens, thus accessing more sophisticated tasks.



%% file: Sections/6_ExtractiveSummarisation.tex

We applied all the special-purpose NLP tools described in Subsection \ref{subsec:nlptools} and the generative models from Subsection \ref{subsec:llms} to the task of extractive summarization.

\subsection{Extractive Summarisation through Special-Purpose NLP Tools}

All special-purpose NLP tools were applied to judicial decisions in order to obtain summaries to be evaluated by legal experts.
Unfortunately, the results  were far from satisfactory (see Section \ref{sec:evaluation}).
In general, we obtained lengthy summaries, in which the relevant legal information was, on the one hand, scattered across the different paragraphs of the summary and, on the other hand, incomplete.

In the following, we provide an English translation of a sentence selected by all the examined techniques:

\begin{example}
Even after this legislative innovation and, therefore, in relation to the current text, the prevailing jurisprudence of the Supreme Court (see lastly Cass. Civ. No. 20981/2021) has adhered to the interpretative option according to which the mere ownership of a real estate asset is not an obstacle to the recognition of the concession, which is instead due to the taxpayer who does not own a property that can be used as a dwelling (in this sense also Cass., Sec. 5, Order No. 19989 of 27/07/2018, according to which ``on the subject of tax concessions for the first home, pursuant to art. 1, note II bis, of the tariff attached to the d.p.r. no. 131 of 1986, in the text (applicable ``ratione temporis") amended by art. 3, paragraph 131, of Law no. 549 of 1995, the concept of ``suitability" of the pre-owned house -- an obstacle to the enjoyment of the benefit (and expressly provided for in the previous legislation) -- must be considered intrinsic to the notion of ``dwelling house" itself, to be understood as accommodation that is concretely suitable, both from an objective-material and legal point of view, to meet the housing needs of the interested party"; as well as Cass., Sec. 5, Judgment No. 2565 of 02/02/2018, which ruled that ``on the subject of first home concessions ... ``the suitability" of the pre-owned dwelling must be assessed both from an objective point of view -- actual uninhabitability -- and from a subjective point of view -- building inadequate in size or qualitative characteristics --, in the sense that the benefit also applies in the case of the availability of accommodation that is not concretely suitable, in terms of size and overall characteristics, to meet the housing needs of the interested party. `` and in the same sense also Cass., Sec. 6-5, Order No. 5051 of 24/02/2021, Cass., Sec. 6-5, Order No. 18091 of 05/07/2019, and Cass., Sec. 6-5, Order No. 18092 of 05/07/2019). 
\end{example}

\subsection{Extractive Summarisation through LLMs} 

Although generative models are not intended to be used for extractive tasks, we attempted to use them for this purpose, too. To this end, we designed a prompt meant to push  generative models toward literal extraction, and we tested it with both GPT-3 and GPT-4. In a preliminary phase, IT5 was also tried, but the results were unsatisfactory, since the model appeared to be unable to follow the instructions for creating the summaries.


Instructions in our prompt were divided into three sections:
\begin{itemize}
\item In the first section, the concept of an ``extractive summary'' goal was declared and defined.
\item In the second section, we described a method for obtaining a summary, i.e., breaking up the document into sentences, assigning a score to each sentence according to  its semantic relevance, and identifying the most relevant sentences.
\item A final section indicated the format in which the output had to be produced, i.e., as a list of sentences and associated scores.
\end{itemize}
We fed out the prompt to GPT 3.5 and 4. Here below we can see the prompt we used and its translation into English.




\noindent
\begin{minipage}[t]{0.48\textwidth}
\begin{lstlisting}
Fra graffe e' presente il contenuto di una sentenza.
Voglio costruire un riassunto estrattivo, cioe' un tasto composto da un sottoinsieme delle frasi presenti nel testo in input.
Per fare un riassunto estrattivo dovresti prima suddividere il testo in frasi, valutare l'importanza semantica, classificarle in base a questa importanza e selezionare le K frasi piu' rilevanti.
In questo caso, K e' uguale a 5.
Le frasi devono essere presentate nello stesso ordine in cui appaiono nel testo originale secondo la seguente struttura.
[1] Frase 1 [Punteggio 1]
...
[K] Frase 1 [Punteggio K]

{ ... }
\end{lstlisting}
\end{minipage}
\hfill
\begin{minipage}[t]{0.48\textwidth}
\begin{lstlisting}
Within the brackets there is the content of a decision.
I want to build an extractive summary, that is a text made of a subset of th phrases i the input text.
To make an extractive summary we should first split the text into sentences, evaluate their semantic importance, and classify them according to this importance and select the K most relevant phrases.
In this case, K is equal to 5.
Phrases must be presented in the same order in which they appear in the original text according to the following structure.
[1] Phrase 1 [Score 1]
...
[K] Phrase 1 [Score K]

{ ... }
\end{lstlisting}
\end{minipage}


The extractive summary obtained through this prompt was shorter than the outcome provided by the specific NLP tools. Following is the output produced feeding GPT-4 with the grounds of the decision in Subsection \ref{rexample}.

\begin{example}
The fundamental preliminary question concerns the evaluation of the suitability or otherwise of the property pre-purchased by the appellant, with respect to the housing needs of himself and his family. It appears from the same that the property in question consists of a small room and a bathroom (depicted in the attached photographic documentation), having a total cadastral area of 25.00 sqm and a usable living area of 17.90 sqm (as shown in the attached plan). The technician also ascertained that the property shows traces of rising damp, mold and fungi, concluding that it is not suitable for meeting the most basic housing needs for a family unit composed of a father and two school-aged children. Based on the aforementioned assessment, supported by plans and photographs, it must therefore be considered established that the property purchased by the respondent\_1 on 19/3/2003, must be considered unsuitable for meeting the housing needs of the aforementioned and his family, so that the first judge correctly considered such purchase not preclusive with respect to the tax benefits claimed. Therefore, the appeal filed must be rejected and the contested judgment must be confirmed
\end{example}

Note that in many cases, unlike the extractive tools, GPT constructed the extractive summaries by selecting and combining phrases or sentence fragments (see, for example, the summary produced by GPT-3 using the same decision in the appendix). In this way, shorter and more informative summaries could be obtained.

%% file: Sections/7_AbstractiveSummarisation.tex

We relied on IT5 and on both versions of GPT to produce abstractive summaries of two kinds: (1) ``flowing-text'' summaries and (2) ``issues-based'' summaries.

\subsection{Flowing-Text Summaries}

Flowing-text abstractive summaries have no prescribed structure. We tested IT5, GPT3, and GPT4 to generate such summaries of each of the two sections of our decisions: ``development of the proceedings'' (\textit{svolgimento del processo}) and ``reasons of the decision'' (\textit{motivi della decisione}). 
For this purpose, a very simple prompt, which we used for all of the three models,  was enough:

\noindent
\begin{minipage}[t]{0.48\textwidth}
\begin{lstlisting}
Fai un sommario del seguente testo tra parentesi graffe

{ ... }
\end{lstlisting}
\end{minipage}
\hfill
\begin{minipage}[t]{0.48\textwidth}
\begin{lstlisting}
Make a summary of the following text within brackets

{ ... }
\end{lstlisting}
\end{minipage}

    

Here below is the summary generated for the decision in Section XXX (development of the proceedings) using GPT4 (the GPT models provide a much better outcome than IT5, with regard to both readability and completeness; see Appendix).

\begin{example}
The text concerns the issue of the suitability of a property pre-purchased by
the applicant to meet the housing needs of his family. The sworn appraisal
filed by the respondent highlights that the property, small in size and with
traces of humidity, mould, and fungi, is not suitable for accommodating a
family unit consisting of a father and two school-aged children. Therefore,
the judge deemed it appropriate not to consider such a purchase as preclu-
sive for the tax benefits claimed. The prevailing case law maintains that
mere ownership of a property is not enough to deny the relief, which is
instead granted to those who do not own a suitable dwelling. The appeal
submitted is rejected and the contested judgment is confirmed, with the
costs of this phase charged to the office.
\end{example}

\subsection{Issue-Based Summaries}\label{sec:IssueBased}

We exploited the capacities of GPT for providing issue-based summaries. The idea is to distinguish the issues addressed by the judges and to provide a separate summary analysis for each of them. This approach was motivated by the hypothesis that this style of summarization may facilitate lawyers in identifying and examining the aspects of the case that are relevant to them. This hypothesis was confirmed by the expert evaluations, which were most favourable for this kind of summary (see Section \ref{sec:evaluation}). 

GPT-3.5 and GPT-4 were instructed on what to look for and list as output by a suitably designed prompt.  We also tested IT5, but this experiment was unsuccessful, since IT5 appeared to be unable to follow the instructions for creating issue-based summaries. 


The set of instructions we used for GPT is devoted to the description of the intended output in two directions: a formal requirement and  a conceptual requirement.

According to the formal requirement, the output consists of a sequence of questions/answer pairs where questions are denoted as QD1, ..., QDn, and the answers are denoted as  PD1, ..., PDn. The prompt also makes it possible to switch between a more human-readable list and a {\tt json} structure.

According to the conceptual requirement, the answers consist in the specification of principles, where a principle is defined as the application or interpretation of an explicit norm, regulation, or previous decision. 

After introducing {\em principles}, the prompt continues by stating that a {\em question} is something that is answered by means of a {\em principle}.

To make the model focus on essential but independent issues, we added a few prescriptions, namely, two {\em principles}  must be very different from each other; the number of {\em principles} in a text is usually 1 or 2 with more {\em principles} appearing only in lengthy texts. 

We also specified that a {\em principles} are to be reported explicitly, and that {\em questions} are to be stated in general terms, i.e., without reference to the specific case at hand.


The adopted prompt is shown below, in the original Italian version and in an English translation (the variant requesting the output in a JSON structure is omitted): 

\noindent
\begin{minipage}[t]{0.48\textwidth}
\begin{lstlisting}
Elenca in una lista nel formato

QD1: testo
PD1: testo

QD2: testo
PD2: testo

...

QDn: testo
PDn: testo

i principi di diritto (PD) e le questioni diritto (QD).

Le QD sono le domande a cui i PD rispondono.
Le QD non contengono nessun riferimento al caso di specie e agli attori della vicenda.

I PD sono le interpretazioni di una o piu norme contenute nel testo tra parentesi graffe.
Per ogni PD specifica i riferimenti alle norme.
Il numero di PD in un testo e di solito 1 o 2.
I PD non contengono nessun riferimento al caso di specie e agli attori della vicenda.
Due PD devono essere molto diversi tra di loro.
In testi lunghi il numero di PD puo essere maggiore di 2.

{ ... }
\end{lstlisting}
\end{minipage}
\hfill
\begin{minipage}[t]{0.48\textwidth}
\begin{lstlisting}
List using the format

QD1: text
PD1: text

QD2: text
PD2: text

...

QDn: text
PDn: text

the legal principles (PD) and legal questions (QD).

QDs are questions answered by PDs.
QDs do not contain any reference to the concrete case and to the involved actors.

PDs are the interpretation of one or more norms contained in the text within brackets.
For every PD specify references to norms.
The number of PDs in a decision is usually 1 or 2.
PDs do not contain any reference to the concrete case and to the involved actors.
Two PDs must be very different from each other.
In long texts, the number of PDs may be greater than 2.

{ ... }
\end{lstlisting}
\end{minipage}

In the following, you can find one of the principles extracted using GPT4:

\begin{example}
\textbf{QD2}: What is the current interpretation of the legislation on tax concessions on a first home in relation to the suitability of a pre-owned dwelling? 

\textbf{PD2}: The suitability of the pre-owned dwelling must be assessed both from an objective point of view (actual uninhabitability) and from a subjective one (inadequate building in terms of size or qualitative characteristics), meaning that the benefit also applies in the case of the availability of a dwelling that is not concretely suitable, in terms of size and overall characteristics, to meet the housing needs of the interested party (Cass., sect. 5, order n. 19989 of 27/07/2018, Cass., sect. 5, judgment n. 2565 of 02/02/2018).
\end{example}

We also experimented with expanding the prompt to include instructions for identifying the original text addressing the summarised issue, and for extracting  keywords.

Line 14 of the original prompt was substituted with

\noindent
\begin{minipage}[t]{0.48\textwidth}
\begin{lstlisting}
i principi di diritto (PD), le questioni diritto (QD), le parole chiave (KW) e le basi testuali (BT).

\end{lstlisting}
\end{minipage}
\hfill
\begin{minipage}[t]{0.48\textwidth}
\begin{lstlisting}
the legal principles (PD), legal questions (QD). keywords (KW) and base texts (BT).

\end{lstlisting}
\end{minipage}

While Line 25 was expanded into

\noindent
\begin{minipage}[t]{0.48\textwidth}
\begin{lstlisting}
Le BT sono le porzioni del testo fra parentesi graffe piu rilevenati per l estrazione di un PD e di una QD.
Le BT non devono contenere variazioni rispetto al testo fra parentesi graffe.
Per ogni QD e PD, restituisci massimo tre BT.

Le KW identificano i temi fondamentali del testo fra parentesi graffe, cioe i concetti giuridici impiegati, gli oggetti disciplinati e le materie trattate.
\end{lstlisting}
\end{minipage}
\hfill
\begin{minipage}[t]{0.48\textwidth}
\begin{lstlisting}
BTs are the portions of the text in brackets most relevant for the extraction of a PD and a QD.
BTs must not contain variations from the text in brackets.
For each QD and PD, return a maximum of three BTs.

KWs identify the fundamental themes of the text in brackets, i.e. the legal concepts used, the objects regulated and the subjects dealt with.
\end{lstlisting}
\end{minipage}




A keyword (KW) refers to relevant legal concepts and subjects contained in the text. The keyword-related portion of the prompt follows the extraction of legal principles, as they should provide contextual information to narrow down the generation to the most relevant keyword with substantive legal meaning.

The prompt also includes instructions for connecting the extracted principles to the relevant part of the original decisions, from which they were extracted (BT). Such a reference allows us to easily verify the correspondence between the extracted principles and the original text. The model is requested to extract three fragments at most, without the introduction of any variation relative to the original text.

Following are the text and keywords associated with the principles in the example above:

\begin{example}
\textbf{BT1}: [on the subject of tax concessions for the first home, pursuant to art. 1, note ii bis, of the tariff attached to the d.p.r. n. 131 of 1986, in the text (applicable ``ratione temporis") amended by art. 3, paragraph 131, of the law n. 549 of 1995]
\textbf{BT2}: [the concept of ``suitability" of the pre-owned home - an obstacle to the enjoyment of the benefit (and expressly provided for in the previous legislation) - must be considered intrinsic to the very notion of ``house of residence", to be understood as a dwelling concretely suitable, both from an objective-material and legal point of view, to meet the housing needs of the interested party]
\textbf{BT3}: [the concessions under examination respond to the reasonable rationale of favoring the purchase of a dwelling in the place of residence or work for the benefit of those who do not have possession of another house of residence objectively suitable to meet their needs]

\textbf{KW}: [tax concessions, first home, housing suitability, housing needs, uninhabitability, inadequacy, property ownership, legislation, jurisprudence]

\end{example}

%% file: Sections/8_Evaluation.tex

The standard automated tools for accessing the quality of summaries, such as ROUGE, do not apply satisfactorily to abstractive summarisation. Thus we submitted questionnaires to tax law experts, asking them to evaluate the process. The questionnaires were previously submitted to the ethical committee of the PRODIGIT project, which reviewed them, proposed refinements and clarifications on the questions and the methodology, and finally approved the revised questionnaires. Based on the indications of the ethical committee, the evaluation only concerned the comparison between different automated systems, to the exclusion of  summaries written by humans. This limitation responds to the following considerations:  on the one hand, having human-prepared summaries for all tax law decisions is not a viable option, this due to the size of that case base;  on the other hand, on the proposed approach,   automated summarisation can be combined with human intervention at the validation and revision stage.

We performed two different evaluations: The first one was devoted to choosing a subset of models that were performing better, this in order to reduce the number of models to be considered. The second one was focused on the resulting small set of models, aiming at collecting from domain experts how these models perform in the context of tax law.

The questionnaires encompassed various evaluation criteria, including:
\begin{itemize}
    \item \textbf{Satisfaction}: Degree of satisfaction with the overall quality of the summary.
    \item \textbf{Correctness}: Accuracy in capturing the source documents' key points, legal nuances, and essential information.
    \item \textbf{Form}: Coherence, readability, and adherence to legal writing conventions.
    \item \textbf{Completeness}: Coverage of important details and comprehensive representation of the source content.
\end{itemize}


The domain experts reviewed the summaries, rated them under each criterion, and provided corresponding scores in the $[1,5]$ range. They also had the option of expressing their comments. They were encouraged to provide their insights, suggestions, and concerns regarding the quality of the summaries. The evaluations were blind, in the sense that the reviewers were not told by what models the summaries were generated. 

By incorporating the feedback and ratings from domain experts through the questionnaires, we obtained a holistic evaluation of the generated summaries. This evaluation process enabled us to gauge the strengths and weaknesses of each model, identify areas for improvement, and make informed decisions about the quality and effectiveness of the generated summaries in accordance with the domain-specific requirements. 


In the next sections, we report and describe the evaluation for the  summaries. 

We report the average scores for each metric considered in Figures \ref{fig:exc_eval}, \ref{fig:abs_eval}, and \ref{fig:sec_eval}. Each graph corresponds to a model, and for each model, we can notice the average value for each metric.

\subsection{First Evaluation}
As observed, the first evaluation had the purpose of selecting the most promising approaches, which would then be subject to the second evaluation. It was carried out by a limited number (12)  of experts in tax law. 

\begin{figure}[htb]
    \centering
    \includegraphics[width=0.8\textwidth]{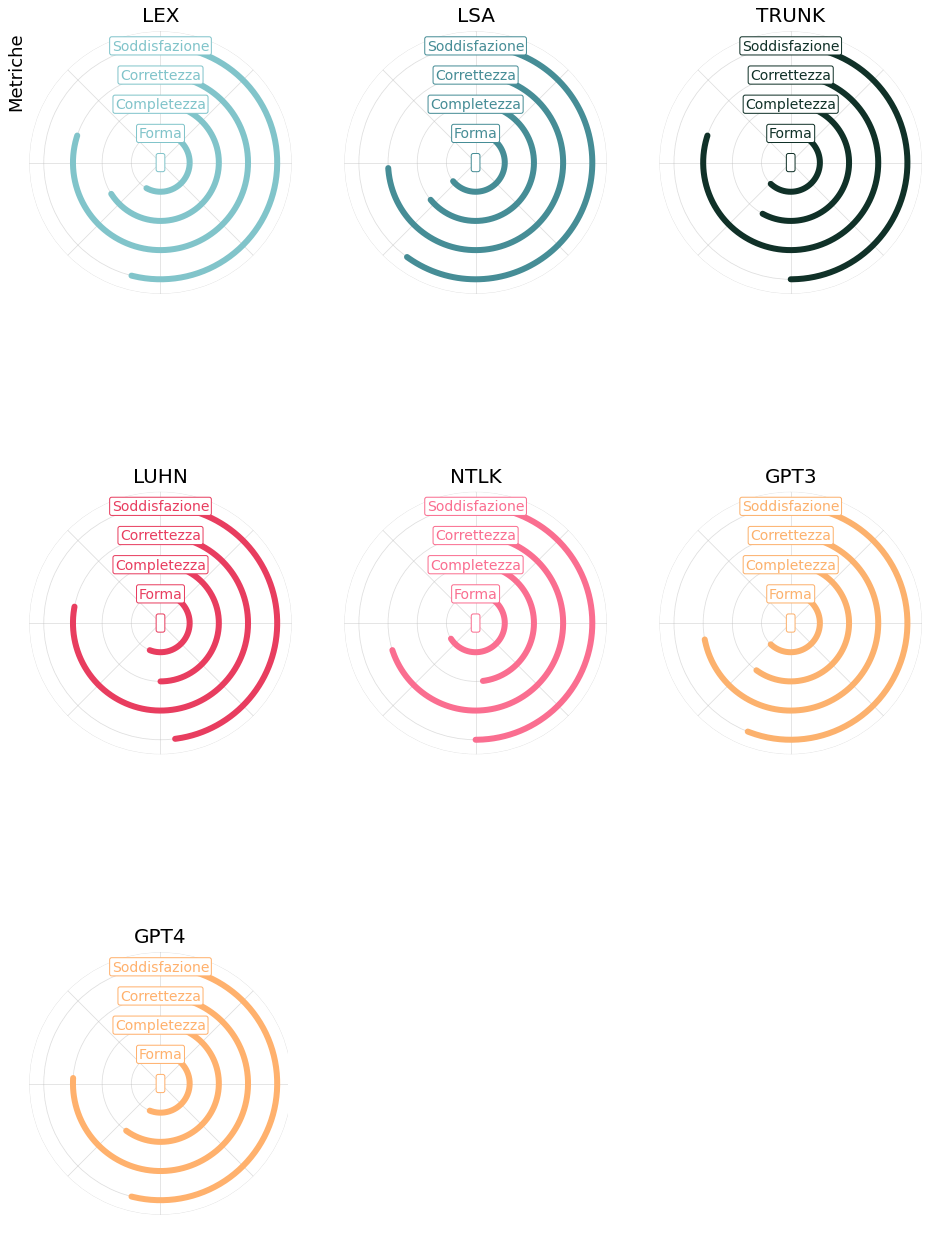}
    \caption{Evaluation score for extractive summaries}
    \label{fig:exc_eval}
\end{figure}

\paragraph{Extractive Summaries}Figure \ref{fig:exc_eval} and Table \ref{tab:first_abs} show that  the special-purpose NLP tools perform  similarly. This is because the extractive nature of the task does not allow for outputs that differ too much in form. The correctness of all the models is quite high, as is to be expected, since they consisted of phrases literally extracted from the original text. Completeness is weak, as the information from other sentences was completely omitted. As a consequence the  the experts' satisfaction was on average low. Similar --through slightly better -- outcomes can  be seen for the extractive summaries produced with generative tools (GPT3 and GPT4).These models as well appear to have omitted much relevant information, especially when dealing with long decisions. On the basis of this evaluation, we decided to discard all extractive approaches and limit the second evaluation to the abstractive methods. However, as shown in Section \ref{sec:IssueBased} we incorporated an extractive aspect in the issue-based summary so as to enable the user to link the abstracted principles to their textual basis.

\begin{table}[htb]
\footnotesize
\begin{tabular}{rccccccc}
 & \textbf{LEX} & \textbf{LSA} & \textbf{TRUNK} & \textbf{LUHN} & \textbf{NTLK} & \textbf{GPT4} & \textbf{GPT3}\\
\toprule
 Form &           2.85          (1.23) &          3.00          (1.11) &          3.09          (1.00) &          2.73          (0.96) &          3.30          (0.90) &          3.10          (0.54) &          2.80          (0.75) \\          
 Completeness &           3.15          (1.51) &          2.85          (1.29) &          2.82          (1.27) &          2.45          (0.99) &          2.40          (1.43) &          3.00          (0.89) &          3.00          (1.10) \\          
 Correctness &           3.69          (1.26) &          3.54          (1.39) &          3.91          (1.00) &          3.73          (1.14) &          3.50          (1.20) &          3.60          (1.11) &          3.80          (1.17) \\          
 Satisfaction &           2.69          (1.32) &          2.77          (1.19) &          2.45          (1.08) &          2.36          (0.88) &          2.50          (1.36) &          2.80          (0.87) &          2.70          (0.64) \\          
\bottomrule
\end{tabular}
\caption{Evaluation of domain experts in the first round of questionnaires on extractive summaries. Standard deviation in brackets.}
\label{tab:first_abs}
\end{table}

\paragraph{Abstractive Summaries}

\begin{figure}[htb]
    \centering
    \includegraphics[width=0.8\textwidth]{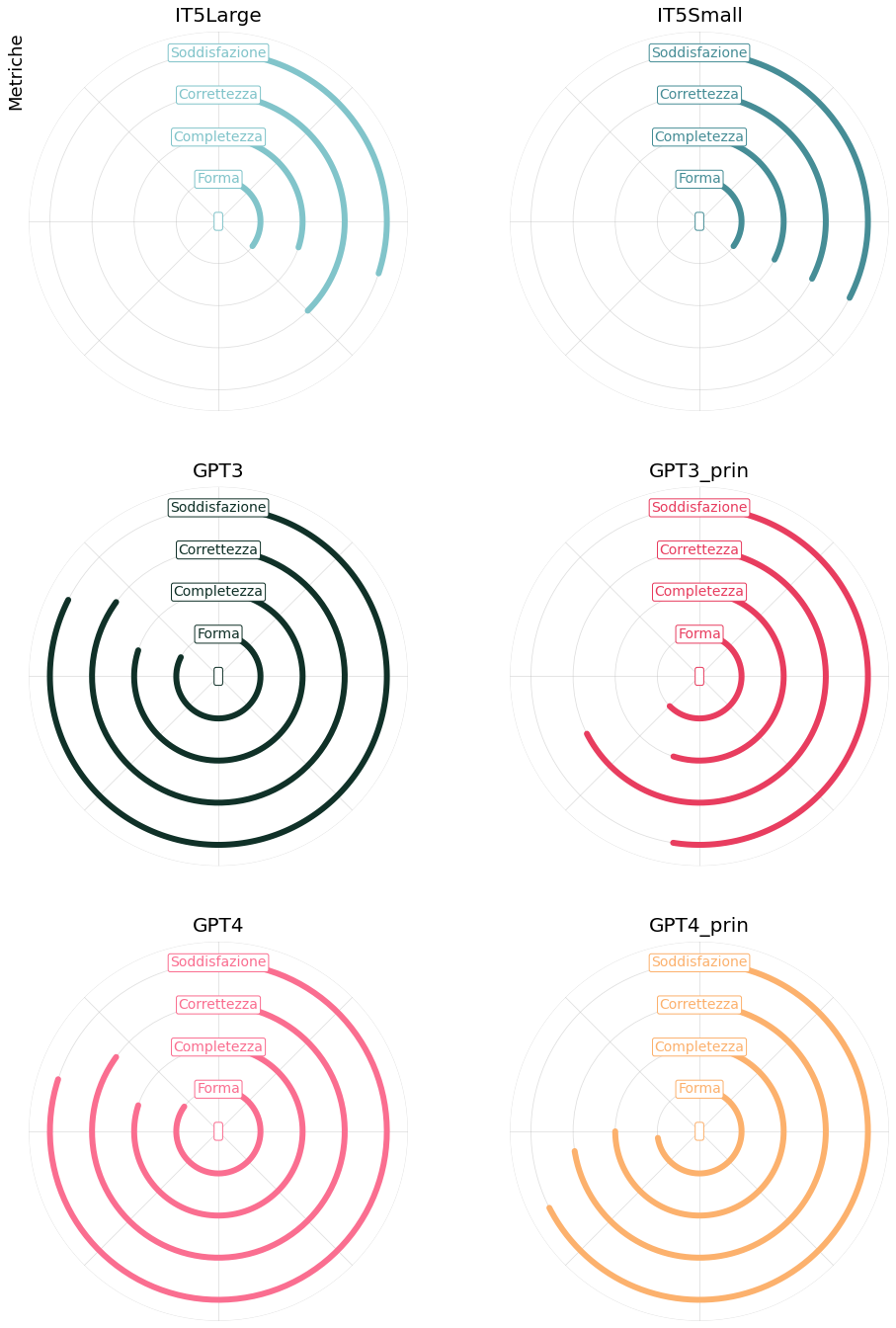}
    \caption{Evaluation score for abstractive summaries (first round)}
    \label{fig:abs_eval}
\end{figure}

In Figure \ref{fig:abs_eval} and Table \ref{tab:first_exc}, we report the average scores for abstractive summaries. 
In this case, IT5 models, which are the baseline, were evaluated as quite poor,  while GPT3 and GPT4 performed very well in all the dimensions. On the basis of this evaluation, we decided for the second stage to keep GPT4 summaries, both in  the flowing text and issue-based versions. We also kept IT5 as a baseline. We chose to omit GPT3, given its similarity to its successor, namely, GPT4.

\begin{table}[htb]
\footnotesize
\begin{tabular}{rcccccc}
 & \textbf{IT5Small} & \textbf{IT5Large} & \textbf{GPT3} & \textbf{GPT4} & \textbf{GPT3 item} & \textbf{GPT4 item}\\
\toprule
 Form &           1.75          (0.97) &          1.75          (1.30) &          4.12          (0.60) &          4.25          (0.66) &          3.12          (1.36) &          3.62          (1.32) \\          
 Completeness &           1.62          (1.32) &          1.50          (1.00) &          4.00          (0.71) &          4.00          (0.71) &          2.75          (1.20) &          3.75          (1.39) \\          
 Correctness &           1.62          (0.99) &          1.88          (1.54) &          4.25          (0.66) &          4.25          (0.66) &          3.38          (1.32) &          3.62          (1.41) \\          
 Satisfaction &           1.62          (1.32) &          1.50          (1.00) &          4.12          (0.33) &          4.00          (0.71) &          2.62          (1.11) &          3.38          (1.32) \\        
\bottomrule
\end{tabular}
\caption{Evaluation of domain experts in the  of questionnaires on abstractive summaries (first round). Standard deviation in brackets.}
\label{tab:first_exc}
\end{table}

\subsection{Second Evaluation:  Abstractive Summaries}

The second evaluation was focused on IT5Small, flowing-text GPT4, and issue-based GPT4.  This evaluation involved around 80 experts, drawn from a pool of judges, lawyers, and others. Each evaluator assessed, for 5 decisions, the summaries produced by the 3 models.

\begin{figure}[htb]
    \centering
    \includegraphics[width=\textwidth]{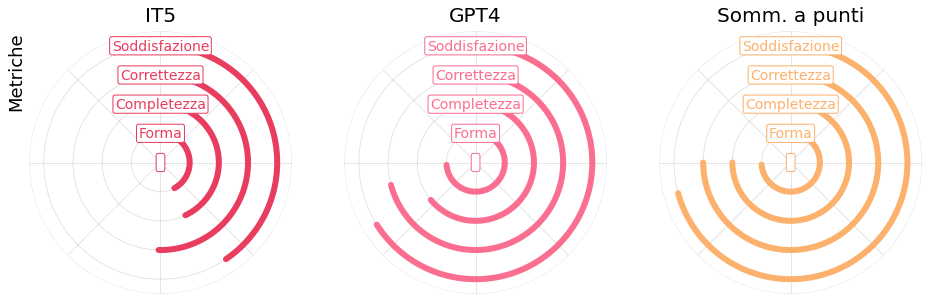}
    \caption{Evaluation score for abstractive summaries (second round)}
    \label{fig:sec_eval}
\end{figure}

We received on average 50 answers for each model. Figure \ref{fig:sec_eval} reports the average values and Table \ref{tab:sec_eval} reports average values and standard deviation.

\begin{table}[htb]
\begin{tabular}{rccc}
& \textbf{IT5} & \textbf{GPT4} & \textbf{GPT4 items} \\
\toprule
Form & 2.11 (1.10) & 3.69 (1.06) & 3.69 (1.14) \\
Completeness & 2.15 (1.06) & 3.20 (1.25) & 3.75 (1.02) \\
Correctness & 2.51 (1.17) & 3.54 (1.15) & 3.75 (1.05) \\
Satisfaction & 2.03 (1.04) & 3.30 (1.32) & 3.54 (1.12)\\
\bottomrule
\end{tabular}
\caption{Evaluation of domain experts in the second round of questionnaires. Standard deviation in brackets.}
\label{tab:sec_eval}
\end{table}

From Figure \ref{fig:sec_eval} it appears that GPT4 scored better than IT5. It also appears that the issue-based summary outperformed the flowing-text summary under both completeness, correctness, and general satisfaction. It can also be seen from Table  \ref{tab:sec_eval} 
  that there is relatively high dispersion in the assessments (according to the indicated standard deviation). We plan to study this aspect to understand whether it relates to the nature of cases, to the language used in the judicial opinions, or to idiosyncrasies of the evaluators.

%% file: Sections/9_RelatedWorks.tex

Since the focus, and the most significant results, of our work concern applying LLMs to the summarization task, the related work most relevant to our project concerns, on the one hand, the use of LLMs in the legal domain and, on the other, automated summarisation.

\subsection{Large Language Models in the Law}

Large language models (LLMs), such as BERT, GPT, or XLM-RoBERTa, have already  demonstrated considerable potential in various legal tasks.

 Notable areas for the deployment of  LLMs have been  judgement prediction and statutory reasoning. The study by \cite{trautmann2022legal} introduces legal prompt engineering (LPE) to enhance LLM performance in tasks involving predicting legal judgements. This method has proven effective across three multilingual datasets, highlighting the model's potential in handling the complexity of legal language and reasoning across multiple sources of information. Another study by \cite{blair2023can} investigates GPT-3’s capacity for statutory reasoning, revealing that dynamic few-shot prompting enables the model to achieve high accuracy and confidence in this task. 

Advancements in prompting techniques have played a crucial
role in the success of LLMs in legal reasoning tasks. The paper by \cite{yu2022legal} introduces Chain-of-Thought (CoT) prompts, which guide LLMs in generating coherent and relevant sentences that follow a logical structure, mimicking a lawyer’s analytical approach. The study demonstrates that CoT prompts outperform baseline prompts in the COLIEE entailment task based on Japanese Civil Code articles.

LLMs have also been employed to understand fiduciary obligations, as explored in \cite{nay2023large}. This study employs natural language prompts derived from US court opinions, illustrating that LLMs can capture the spirit of a directive, thus facilitating more effective communication between AI agents and humans using legal standards. 

The potential of LLMs in legal education has been examined in studies
such as \cite{choi2023chatgpt} and \cite{hargreaves2023words}. In \cite{choi2023chatgpt}, the authors task ChatGPT with writing law school exams without human assistance, revealing potential concerns and insights about LLM capabilities in legal assessment.
On the other hand, the paper by \cite{hargreaves2023words} addresses the ethical use of AI language models like ChatGPT in law school assessments, proposing ways to teach students appropriate and ethical AI usage. 

The role of LLMs in supporting law professors and providing legal advice has also been investigated. The study in \cite{oltz2023chatgpt} suggests that LLMs can assist law professors in administrative tasks and streamline scholarly activities. 

Furthermore, LLMs have been explored as quasi-expert legal advice lawyers in \cite{macey2023chatgpt}, showcasing the possibility of using AI models to support individuals seeking affordable and prompt
legal advice. The potential impact of LLMs on the legal profession has been a subject of debate, as discussed in \cite{iu2023chatgpt}. This paper evaluates the extent to which ChatGPT can serve as a replacement for
litigation lawyers by examining its drafting and research capabilities. 

Finally, the study by \cite{nay2022law} proposes a legal informatics approach to align AI with human goals and societal values. By embedding legal knowledge and reasoning in AI, the paper contributes to the research agenda of integrating AI
and law more effectively.
In conclusion, LLMs have shown promising results in various legal tasks, with the advancement of prompting techniques playing a crucial role in their success. However, challenges remain in ensuring
the ethical use of LLMs and addressing their potential impact on the legal profession.

\subsection{Legal Text Summarization}

Summarisation of has been   a forefront task in legal informatics for some years. In 2004 a seminal contribution \cite{grover2004holj} provided the extractive summarisation of  a legal dataset of 188 judgements from the House of Lords Judgement (HOLJ) website from 2001–2003. However, only recently have researchers started to produce promising results,  thanks to state-of-the-art NLP, machine learning techniques, and, lately, LLMs.

Existing research on legal summarization mostly applies  extractive methods. A wide range of approaches to this effect exist, from  classical algorithms~\cite{Gong2001, Erkan2004, Brin1998} to domain-specific methods. Among the latter, there are works based on nature-inspired methods, i.e., algorithms emulating natural processes~\cite{Yang2014}, using optimization approaches that adapt to challenging circumstances~\cite{Kanapala2019}; graph-based methods, where sentences are selected based on the construction and search over similarity graphs \cite{Erkan2004,Kim2013,Duan2019}; and citation-based methods relying upon the set of citing sentences within documents to build summaries~\cite{Galgani2015}. Finally, there are also machine-learning-based models in which classifiers predict which sentences to include in the summary \cite{Zhong2019}.  

Kanapala et al. \cite{Kanapala2019}   focused on a domain-specific  automatic summarization system based on a nature-inspired method. The authors framed legal document summarization as a binary optimization problem, utilizing statistical features such as sentence length, position, similarity, term frequency-inverse sentence frequency, and keywords.  The authors used the gravitational search algorithm (GSA) as the optimization technique for generating summaries. GSA adjusted the weights assigned to sentence features, capturing the importance and relevance of sentences within legal documents. To evaluate their method, the authors compared it with other approaches, including genetic algorithms, particle swarm optimization, TextRank, latent semantic analysis (LSA), MEAD, SumBasic, and MS-Word summarizer. They utilized the FIRE-2014 dataset, which consisted of 1,000 Supreme Court judgements from 1950 to 1989. The proposed algorithm outperformed the other methods based on ROUGE evaluation metrics.

Merchant and Pande \cite{Merchant2018} presented an automated text summarization system designed to help lawyers and citizens conduct comprehensive research for their legal cases. The researchers used LSA, a natural language processing technique, to capture concepts within individual documents. Two approaches were used -- a single-document untrained approach and a multi-document trained approach -- depending on the type of case (criminal or civil). The data used in the study was collected from the Indian official government websites and included  Supreme Court, High Court, and District Court cases. The evaluation of the model resulted in an average ROGUE-1 score of 0.58. The system received the approval of professional lawyers.

Licari et al. \cite{Licari2023} introduce a method for automatically extracting legal holdings from Italian cases using Italian-LEGAL-BERT and present a benchmark dataset called ITA-CaseHold for Italian legal summarization. They introduced HM-BERT, an extractive summarization tool based on Italian-LEGAL-BERT. HM-BERT selects relevant sentences using a similarity function based on unigram and bigram overlap. The model achieved prominent results in terms of ROUGE scores, and the extracted holdings were validated by experts. The paper acknowledges limitations such as potential redundancy in sentence selection and the challenge of explaining HM-BERT's decisions.

Recently, in connection with the availability of transformer models, some attempts at  abstractive summarisation have been developed. 
Schraagen et al. \cite{schraagen-etal-2022} applied two abstractive models to a Dutch legal domain dataset and evaluated their performance using ROUGE scores and evaluation by legal experts. The study presents a hybrid model based on reinforcement learning and a transformer-based BART model trained on a large dataset of Dutch court judgements. The results show promising transferability of the models across domains and languages, with ROUGE scores comparable to state-of-the-art studies on English news articles. However, human evaluation shows that handwritten summaries are still perceived as more relevant and readable. Furthermore, summarisers struggle to include all necessary elements in the summary, leading to the omission of important details. The authors suggest that  the abstractive summarisation process can be improved by incorporating domain-specific constraints,  such as focusing on citations of legal sources and structuring summaries into facts, arguments, and decisions.

Prabhakar et al.~\cite{Prabhakar2022} presented a method using  T5  to generate abstractive summaries of Indian legal judgements. The system uses a dataset of 350 judgements of the Honourable Supreme Court of India,  compiled with the assistance of a lawyer. The generated summaries are evaluated using the ROUGE score, with a Rouge-L precision of $0.54955$. 

Feijo et al.~\cite{Feijo2023} addressed the problem of ``hallucination'' in abstractive text summarisation, focusing specifically on legal texts. They proposed a novel method, called \textit{LegalSumm}, which aimed to improve the fidelity and accuracy of the generated summaries. To achieve this, the authors created multiple ``views'' of the source text and trained summarisation models to generate independent versions of the summaries. They also introduced an entailment module to evaluate the fidelity of candidate summaries to the source text. The authors demonstrated the effectiveness of their approach by showing significant improvements in ROUGE scores across all evaluation metrics. As well as contributing to the field of legal summarisation, the study provides a basis for further advances in the production of reliable and accurate summaries.

Koniaris et al.~\cite{Koniaris2023} addressed the challenge of automatic summarization of Greek legal documents. To overcome the lack of suitable datasets in the Greek language, the authors developed a metadata-rich dataset of selected judgements from the Supreme Civil and Criminal Court of Greece, along with their reference summaries and category tags. They adopted state-of-the-art methods for abstractive (BERT) and extractive (LexRank) summarization and conducted a comprehensive evaluation using both human and automatic metrics, such as ROUGE. The results showed that extractive methods had average performance, while abstractive methods generated moderately fluent and coherent text but received low scores in relevance and consistency metrics. They identified the need for better metrics to evaluate legal document summaries' coherence, relevance, and consistency. The authors suggested future research directions involving better datasets and improved evaluation metrics, as well as exploring advanced techniques such as deep learning with various neural network architectures to enhance the quality of generated summaries.

Huang et al.~\cite{Huang2023} proposed a two-stage legal judgement summarization model to address the challenges posed by lengthy legal judgements and their technical terms. They leveraged raw legal judgements with varying granularities as input information and treated them as sequences of sentences. Key sentence sets were selected from the full texts to serve as the input corpus for the summary generation. Additionally, the authors incorporated an attention mechanism by extracting keywords related to technical terms and specific topics in the legal texts, which were integrated into the summary-generation model. Experimental evaluations on the CAIL2020 and LCRD datasets demonstrated that their model, based on recurrent neural networks and attention mechanisms, outperformed baseline models (Lead-3, TextRank, and others), achieving an overall improvement of 0.19–0.41 in ROUGE scores. The results indicated that their method effectively captured essential and relevant information from lengthy legal texts and generated improved legal judgement summaries. 




%% file: Sections/10_Conclusions.tex

We have presented some preliminary results obtained in the early development of the PRODIGIT project, a tax law initiative aimed at supporting  judges, lawyers, and other legal practitioners.  We have focused on the summarisation task, one of the main goals of PRODIGIT, which aims to support the summarisation of all Italian tax law decisions.  

We first  introduced tax law adjudication in the Italian legal system,  discussed the significance of summarisation for judges and practitioners, and described the database we  used for our experiments.
We then introduced the  tools and approaches we used in our experiments, which range from  single special-purpose NLP tools, based on classical statistical approaches, to the most recent LLMs. 
We described our experiments with regard to all such tools, providing examples of the obtained outputs and discussing  the limitations and potentialities of each approach. In particular, we provided an in-depth account of our use of generative LLMs, which yielded clearly superior results.
In this regard, we listed the prompts used to obtain such results. The most interesting approach we developed is that of ``issue-based summarisation'', i.e.,  outlining the  legal issues examined by the judges and the corresponding legal criteria (principles). Issue-based summarisation was complemented by the  extraction of issue-relative keywords and textual fragments. For this purpose an appropriate prompt had to be devised to direct generative tools toward providing legally meaningful information. 
We think that this is the most significant development our work provides in relation to the literature cited in Section~\ref{sec:related-work}. 

We have submitted the results of our experiments to an extensive evaluation by legal experts, from which emerged a clear preference for the outcomes delivered by generative tools. In particular, the issue-based summarisation delivered by GPT4 appeared to be the preferred approach, considering its linguistic quality, completeness, and correctness, as well as general satisfaction.
This comparative assessment is also a significant innovative contribution to the theory and practice of legal summarisation, which cannot avoid facing the challenge of LLMs.

We think that some interesting lessons emerge from our experience.

The first takeaway is that the most advanced LLMs can provide very good results in automated summarisation, clearly outperforming earlier NLP tools. Different kinds of summaries can be obtained by carefully designing the corresponding prompt. We have observed that high-quality outcomes can be obtained even without fine-tuning large LLMs. The extent to which fine-tuning can provide improvements in performance is an important issue for further research.

The second takeaway is that extensive human evaluation is needed to assess the outcomes of summarisation in the legal domain. As noted, we submitted the results of our experiment to expert evaluation, based on questionnaires reviewed by an ethical committee. This evaluation provided us with a clear comparative assessment of the summaries, on which the implementation of summarisation within PRODIGIT will be based. We do not think that, at the state of the art, any automated methods can be deployed to test the quality of summarisation, particularly in the abstractive case. In addition to the two-step formal evaluation described in Section~\ref{sec:evaluation}, in multiple round we submitted our preliminary results to the judgement of expert tax lawyers involved in the project. This enabled us to refine our methods, and in particular to refine our prompts in order to approximate the desired outcomes before the formal evaluation.

A third takeaway is that summarisation provides different satisfactory outcomes with regard to different input texts. In fact, Italian tax law decisions very greatly in  length,  in the language used, and in the clarity of reasoning. In some cases, making sense of their content may be a challenge for human readers; thus it is not surprising that automated summarisers give different results. Thus, it is important to preserve convenient human supervision over the outcomes of automated summarization. Under the PRODIGIT project, an application is being developed to enable expert lawyers to review and possibly revise the automatically generated summaries. Their feedback will support further improvement of automated summarisation. 

Based on the successful  experiments described in this paper, summarisation of the PRODIGIT project will take steps toward putting out publicly available results.

The idea is to include the automatically generated summaries in a publicly accessible database of all  Italian tax law decisions: we first provide the summaries of decisions in the registration-tax area will, and if the public's response is positive, the exercise will be extended to all domains of tax law.

We are also experimenting with using summaries -- in the issue-based versions -- for the purpose of indexing and searching the case law. The extracted information -- legal issues and keywords -- are being  used to construct a conceptual graph through which to access the case-law database.

%% file: Sections/11_Acknowledgments.tex
This work was developed within the scope of the PRO.DI.GI.T project (Project for Innovation in Tax Justice, with the support of digital technologies and AI -- Progetto per l’innovazione della Giustizia Tributaria, con il supporto della tecnologia digitale e dell’intelligenza artificiale) proposed and endorsed by the Presidency Council of Tax Justice (Consiglio di Presidenza della Giustizia Tributaria - CPGT) and 
the Ministry of Economy and Finance (Ministero dell'Economia e delle Finanze - MEF). We gratefully acknowledge SOGEI for its support and cooperation during the project.

%% file: Sections/Appendix.tex
In the following, we present the text related to our running example, namely, Decision No. 7683 of 14 September 2022 issued by the Court of Second Instance of Sicily (see Section \ref{rexample}).
\subsection{Original Text}\label{app:orig}

\paragraph{Development of the Proceedings}

With separate judgments issued by the Provincial Tax Commission of Messina, the appeals filed by respondent\_1 were accepted, respectively against the notice of tax assessment and imposition of sanctions with which the office had determined a higher registration tax of € 15,890.00, plus interest and penalties following the revocation of tax benefits for the purchase of the first home, relating to the award of the property located in place\_1, address\_1, in the land registry at folio 222, part. 87 sub 9; and against the notice of recovery of the substitute tax on the related mortgage transaction (at the ordinary rate of 2\% instead of the reduced rate of 0.25\%). The assessment originated from the finding that the taxpayer had actually previously purchased a separate residential property in the NCEU of place\_1 at fol. 216, part. 115 sub 43, category A/4; so he would not have been in the condition of not being the exclusive owner or in communion with his spouse, of the property rights on another dwelling house. The first judge had annulled the contested acts, starting from the consideration that the ``first home" benefit - governed by article 1 of the tariff - first part, note II bis, attached to the Presidential Decree no. 131/86 - allowed the appellant to benefit from the tax benefit (consisting in the application of the registration tax with the rate of 2\% instead of 9\%), since the appellant was indeed the owner of another real estate unit, purchased with a deed of sale registered in place\_2 on 19.03.2003 at no. 277, but the same property was to be considered unsuitable for the housing needs of the appellant and his family, due to its consistency and intrinsic characteristics. Against these decisions, the second of which was consequential to the first, the office filed an appeal, complaining of their incorrectness both because the pre-acquired property was still classified as a residential category, and because its unsuitability could not be considered based on the simple assertions of the appellant, regarding the alleged size of 18 square meters, contradicted by the attached registry extract. Respondent\_1 appeared in both judgments, contesting the appeal filed, for which he requested dismissal, producing extensive case law together with a sworn technical consultancy of the party. Therefore, at the hearing of 14/9/2022, duly scheduled, the two connected proceedings were joined and at the end, the commission decided as per the device.

\paragraph{Grounds for the Decision}

The acts of grievance are unfounded and must be rejected. A fundamental preliminary question concerns the assessment of the suitability or otherwise of the property pre-purchased by the applicant, with respect to the housing needs of himself and his family. For this purpose, beyond mere assertions and disputes, it is useful to refer to the content of the sworn expert report signed by nominee\_1, filed by the appellant during the appeal phase. It emerges from this report that the property in question consists of a small room and a bathroom (depicted in the attached photographic documentation), with a total cadastral area of 25.00 square meters and a usable living area of 17.90 square meters (as shown in the attached plan). In particular, it appears that the property develops only and exclusively on the front facing the inner courtyard and that the single room has a high vasistas window overlooking the same inner courtyard, while the bathroom lacks a window but is equipped with a forced mechanical ventilator. The technician also ascertained that the property shows traces of rising damp, mold, and fungi, concluding that it is not suitable for meeting the most basic housing needs for a family unit composed of a father and two school-age children. Based on the cited assessment, supported by plans and photographs, it must therefore be considered established that the property purchased by the resisting party\_1 on 19/3/2003, must be considered unsuitable to meet the housing needs of the aforementioned and his family, so that the first judge correctly considered such purchase not preclusive with respect to the fiscal benefits invoked.

A brief regulatory and jurisprudential excursus must be carried out in support: Article 16 of the d.l. 22 May 1993, n. 155, amending the T.U. registration tax, had prescribed, for the recognition of the concession, that in the purchase deed the buyer declared not to possess another building ``suitable for habitation" in the same municipality, but the subsequent law 28 December 1995, n. 549 (with art. 3, paragraph 131) has innovated the regulatory text which, in the version applicable at the time of purchase (and still in force), establishes ``that in the purchase deed the buyer declares not to be the exclusive holder or in communion with the spouse of the rights of property, usufruct, use and habitation of another dwelling house in the territory of the municipality in which the property to be purchased is located". Even after this legislative innovation and, therefore, in relation to the current text, the prevailing jurisprudence of the Supreme Court (see lastly Cass. Civ. n. 20981/2021) has adhered to the interpretative option according to which the mere ownership of a real estate asset is not an obstacle to the recognition of the concession, which is instead due to the taxpayer who does not own a property that can be used for habitation (in this sense also Cass., sect. 5, order no. 19989 of 27/07/2018, according to which ``in the matter of tax concessions for the first home, pursuant to art. 1, note II bis, of the tariff attached to d.p.r. n. 131 of 1986, in the text (applicable ``ratione temporis") amended by art. 3, paragraph 131, of l. n. 549 of 1995, the concept of ``suitability" of the pre-owned house - an obstacle to the enjoyment of the benefit (and expressly provided for in the previous legislation) - must be considered intrinsic to the notion itself of ``dwelling house", to be understood as accommodation concretely suitable, both from an objective-material and legal point of view, to meet the housing needs of the interested party"; as well as Cass., sect. 5, judgment no. 2565 of 02/02/2018, which ruled that ``in the matter of first home concessions ... ``the suitability" of the pre-owned dwelling must be assessed both from an objective point of view - actual uninhabitability - and from a subjective point of view - building inadequate in size or qualitative characteristics - , in the sense that the benefit applies even in the case of the availability of accommodation that is not concretely suitable, in terms of size and overall characteristics, to meet the housing needs of the interested party." and in the same sense also Cass., sect. 6-5, order no. 5051 of 24/02/2021, Cass., sect. 6-5, order no. 18091 of 05/07/2019, and Cass., sect. 6-5, order no. 18092 of 05/07/2019). It is also worth remembering that the same Constitutional Court, in declaring inadmissible the question of constitutional legitimacy of letter b) of number 1) of note II-bis of art. 1 of the first part of the tariff attached to d.p.r. 26 April 1986, n. 131 (raised with reference to articles 3 and 53 of the Constitution and the principles of reasonableness, rationality and non-contradiction), with order no. 203 of 6 July 2011, has expressly recognized that ``the concessions in question respond to the reasonable ratio of favoring the purchase of a dwelling in the place of residence or work for the benefit of those who, in the same place, do not have possession of another dwelling house objectively suitable to meet their needs". 
From this it is further confirmed the correctness of the interpretation according to which the ownership of another property that is unsuitable, due to its small size, to be used as a dwelling is not an obstacle to the application of the ``first home" concessions. Therefore, the appeal proposed must be rejected and the contested judgment must be confirmed. The costs of this phase, following the defeat, must be charged to the office and liquidated as per the device, also considering the reunion of the two original separate proceedings.

\paragraph{Final Ruling}

For this reason, the Court rejects the appeal and confirms the contested judgment. It condemns the appellant to the payment of the legal costs, which are settled at a total of Euro 1,850.00 plus accessories as provided by law.

\subsection{Extractive Summaries of the Reasons for the Decision}\label{app:extr}


\begin{center}
\begin{longtable}{p{0.15\textwidth}|p{0.8\textwidth}}
\caption{Summaries generated by extractive methods.} \label{tab:extractive} \\

\hline \multicolumn{1}{c}{\textbf{Technique}} & \multicolumn{1}{c}{\textbf{Summary}} \\ \hline 
\endfirsthead

\multicolumn{2}{c}%
{{\textbf{\tablename\ \thetable{}} -- Summaries generated by extractive methods (Continued)}} \\
\hline \multicolumn{1}{c}{\textbf{Technique}} & \multicolumn{1}{c}{\textbf{Summary}}  \\ \hline 
\endhead

\multicolumn{2}{r}{{Continued on next page}} \\ \hline
\endfoot

\endlastfoot
\textbf{LSA} & \small The technician also ascertained that the property shows signs of rising damp, mold, and fungi, concluding that it is not suitable for meeting the most basic housing needs for a family unit consisting of a father and two school-age children. Even after this legislative innovation and, therefore, in relation to the current text, the prevailing jurisprudence of the Supreme Court (see lastly Cass. Civ. No. 20981/2021) has adhered to the interpretative option according to which the mere ownership of a real estate asset is not an obstacle to the recognition of the benefit, which is instead due to the taxpayer who does not own a property that can be used as a dwelling (in this sense also Cass., Sec. 5, Order No. 19989 of 27/07/2018, according to which ``in the matter of tax benefits for the first home, pursuant to art. 1, note II bis, of the tariff attached to d.p.r. No. 131 of 1986, in the text (applicable ``ratione temporis") amended by art. 3, paragraph 131, of Law No. 549 of 1995, the concept of ``suitability" of the pre-owned house - an obstacle to the enjoyment of the benefit (and expressly provided for in the previous legislation) - must be considered intrinsic to the notion of ``house of residence" itself, to be understood as a dwelling that is concretely suitable, both from an objective-material and legal point of view, to meet the housing needs of the interested party"; as well as Cass., Sec. 5, Judgment No. 2565 of 02/02/2018, which ruled that ``in the matter of first home benefits ... the ``suitability" of the pre-owned dwelling must be assessed both from an objective point of view - actual uninhabitability - and from a subjective point of view - building inadequate in size or qualitative characteristics -, in the sense that the benefit applies even in the case of availability of a dwelling that is not concretely suitable, in terms of size and overall characteristics, to meet the housing needs of the interested party." In the same sense, Cass., Sec. 6-5, Order No. 5051 of 24/02/2021, Cass., Sec. 6-5, Order No. 18091 of 05/07/2019, and Cass., Sec. 6-5, Order No. 18092 of 05/07/2019 also ruled. It is also worth remembering that the Constitutional Court itself, in declaring inadmissible the question of the constitutional legitimacy of letter b) of number 1) of note II-bis of art. 1 of the first part of the tariff attached to d.p.r. April 26, 1986, No. 131 (raised in reference to Articles 3 and 53 of the Constitution and to the principles of reasonableness, rationality, and non-contradiction), with Order No. 203 of July 6, 2011, expressly recognized that ``the benefits in question respond to the reasonable rationale of favoring the purchase of a dwelling in the place of residence or work for the benefit of those who, in the same place, do not have possession of another house of residence objectively suitable to meet their needs." \\
\midrule
\textbf{LexRank} & \small The registry tax had prescribed, for the recognition of the concession, that in the deed of purchase the buyer declared not to own another building ``suitable for habitation" in the same municipality, but the subsequent law of December 28, 1995, no. 549 (with art. 3, paragraph 131) has innovated the regulatory text which, in the version applicable at the time of purchase (and still in force), establishes ``that in the deed of purchase the buyer declares not to be the exclusive owner or in communion with the spouse of the rights of ownership, usufruct, use and habitation of another dwelling house in the territory of the municipality where the property to be purchased is located." Even after this legislative innovation and, therefore, in relation to the current text, the prevailing jurisprudence of the Supreme Court (see lastly cass. civ. no. 20981/2021) has adhered to the interpretative option according to which the mere ownership of a real estate asset is not an obstacle to the recognition of the concession, which is due, instead, to the taxpayer who does not own a property that can be used as a dwelling (in this sense also cass., sect. 5, order no. 19989 of 27/07/2018, according to which ``in the matter of tax concessions for the first home, pursuant to art. 1, note ii bis, of the tariff attached to d.p.r. no. 131 of 1986, in the text (applicable ``ratione temporis") amended by art. 3, paragraph 131, of l. no. 549 of 1995, the concept of ``suitability" of the pre-owned house - an obstacle to the enjoyment of the benefit (and expressly provided for in the previous legislation) - must be considered intrinsic to the notion of ``dwelling house" itself, to be understood as a dwelling that is concretely suitable, both from an objective-material and legal point of view, to satisfy the housing needs of the interested party"; as well as cass., sect. 5, judgment no. 2565 of 02/02/2018, which ruled that ``in the matter of first home concessions ... ``the suitability" of the pre-owned dwelling must be assessed both from an objective point of view - actual uninhabitability -, and from a subjective point of view - building inadequate in size or qualitative characteristics -, in the sense that the benefit also applies in the case of availability of a dwelling that is not concretely suitable, in terms of size and overall characteristics, to satisfy the housing needs of the interested party." In the same sense, cass., sect. 6-5, order no. 5051 of 24/02/2021, cass., sect. 6-5, order no. 18091 of 05/07/2019, and cass., sect. 6-5, order no. 18092 of 05/07/2019 also ruled. Therefore, the appeal filed must be rejected and the contested judgment must be confirmed. \\
\midrule
\textbf{TextRank} & \small The registry tax had prescribed, for the recognition of the concession, that in the act of purchase the buyer declared not to own another building ``suitable for habitation" in the same municipality, but the subsequent law of December 28, 1995, n. 549 (with art. 3, paragraph 131) has innovated the regulatory text which, in the version applicable at the time of the purchase (and still in force), establishes ``that in the act of purchase the buyer declares not to be the exclusive owner or in communion with the spouse of the rights of ownership, usufruct, use and habitation of another dwelling house in the territory of the municipality in which the property to be purchased is located". Even after this legislative innovation and, therefore, in relation to the current text, the prevailing jurisprudence of the Supreme Court (see lastly Cass. Civ. No. 20981/2021) has adhered to the interpretative option according to which the mere ownership of a real estate asset is not an obstacle to the recognition of the concession, which is instead due to the taxpayer who does not own a property that can be used for habitation (in this sense also Cass., Section 5, order No. 19989 of 27/07/2018, according to which ``in the matter of tax concessions for the first home, pursuant to art. 1, note II bis, of the tariff attached to the d.p.r. No. 131 of 1986, in the text (applicable ``ratione temporis") amended by art. 3, paragraph 131, of Law No. 549 of 1995, the concept of ``suitability" of the pre-owned house - an obstacle to the enjoyment of the benefit (and expressly provided for in the previous legislation) - must be considered intrinsic to the notion of ``dwelling house" itself, to be understood as accommodation concretely suitable, both from an objective-material and legal point of view, to satisfy the housing needs of the interested party"; as well as Cass., Section 5, judgment No. 2565 of 02/02/2018, which ruled that ``in the matter of first home concessions ... ``the suitability" of the pre-owned dwelling must be assessed both from an objective point of view - actual uninhabitability - and from a subjective point of view - building inadequate in size or qualitative characteristics - , in the sense that the benefit applies even in the case of the availability of accommodation that is not concretely suitable, in terms of size and overall characteristics, to meet the housing needs of the interested party." And in the same sense, Cass., Section 6-5, order No. 5051 of 24/02/2021, Cass., Section 6-5, order No. 18091 of 05/07/2019, and Cass., Section 6-5, order No. 18092 of 05/07/2019 also pronounced. It is also worth remembering that the Constitutional Court itself, in declaring inadmissible the question of constitutional legitimacy of letter b) of number 1) of note II-bis of art. 1 of the first part of the tariff attached to the d.p.r. April 26, 1986, No. 131 (raised with reference to articles 3 and 53 of the Constitution and to the principles of reasonableness, rationality and non-contradiction), with order No. 203 of July 6, 2011, expressly recognized that ``the concessions in question respond to the reasonable rationale of promoting the purchase of a dwelling in the place of residence or work for the benefit of those who, in the same place, do not have possession of another dwelling house objectively suitable to meet their needs". \\
\midrule
\textbf{Luhn} & \small It results from the same that the property in question consists of a small room and a bathroom (depicted in the attached photographic documentation), with a total cadastral area of 25.00 square meters and a usable living area of 17.90 square meters (as shown in the attached plan). In particular, it appears that the property develops exclusively on the facade overlooking the inner courtyard, and that the single room is equipped with a high vasistas window overlooking the same inner courtyard, while the bathroom has no window but is equipped with a forced mechanical ventilator. Even after this legislative innovation and, therefore, in relation to the current text, the prevailing jurisprudence of the supreme college (see lastly cass.civ n 20981/2021) has adhered to the interpretative option according to which the mere ownership of a real estate asset does not preclude the recognition of the benefit, which is instead granted to the taxpayer who does not own a property that can be used as a dwelling (in this sense also cass, sec 5, order n 19989 of 27/07/2018, according to which ``in the matter of tax benefits for the first house, pursuant to art 1, note ii bis, of the tariff attached to dpr n 131 of 1986, in the text (applicable ``ratione temporis") amended by art. 3, paragraph 131, of the l. n 549 of 1995, the concept of ``suitability" of the pre-owned house - an obstacle to the enjoyment of the benefit (and expressly provided for in the previous legislation) - must be considered intrinsic to the notion of ``dwelling house" itself, to be understood as accommodation that is concretely suitable, both from an objective-material and legal point of view, to meet the housing needs of the person concerned"; as well as cass, sec 5, judgment n 2565 of 02/02/2018, which ruled that ``in the matter of first home benefits ... ``the suitability" of the pre-owned dwelling must be assessed both from an objective point of view - actual uninhabitability - and from a subjective point of view - building inadequate in size or qualitative characteristics - in the sense that the benefit is also applicable in the case of availability of accommodation that is not concretely suitable, in terms of size and overall characteristics, to meet the housing needs of the person concerned." and in the same sense, cass, sec 6-5, order n 5051 of 24/02/2021, cass, sec 6-5, order n 18091 of 05/07/2019, and cass, sec 6-5, order n 18092 of 05/07/2019 have also been pronounced). \\
\midrule
\textbf{NLTK} & \small The registry tax had prescribed, for the recognition of the concession, that in the deed of purchase the buyer declared not to own another building ``suitable for habitation" in the same municipality, but the subsequent law of December 28, 1995, no. 549 (with art. 3, paragraph 131) has innovated the regulatory text which, in the version applicable at the time of purchase (and still in force), establishes ``that in the deed of purchase the buyer declares not to be the exclusive owner or in communion with the spouse of the rights of ownership, usufruct, use and habitation of another dwelling house in the territory of the municipality where the property to be purchased is located". Even after this legislative innovation and, therefore, in relation to the current text, the prevailing jurisprudence of the Supreme Court (see lastly Cass. Civ. No. 20981/2021) has adhered to the interpretative option according to which the mere ownership of a real estate asset is not an obstacle to the recognition of the concession, which is instead due to the taxpayer who does not own a property that can be used as a dwelling (in this sense also Cass., Sec. 5, Order No. 19989 of 27/07/2018, according to which ``on the subject of tax concessions for the first home, pursuant to art. 1, note II bis, of the tariff attached to the d.p.r. no. 131 of 1986, in the text (applicable ``ratione temporis") amended by art. 3, paragraph 131, of Law no. 549 of 1995, the concept of ``suitability" of the pre-owned house - an obstacle to the enjoyment of the benefit (and expressly provided for in the previous legislation) - must be considered intrinsic to the notion of ``dwelling house" itself, to be understood as accommodation that is concretely suitable, both from an objective-material and legal point of view, to meet the housing needs of the interested party"; as well as Cass., Sec. 5, Judgment No. 2565 of 02/02/2018, which ruled that ``on the subject of first home concessions ... ``the suitability" of the pre-owned dwelling must be assessed both from an objective point of view - actual uninhabitability - and from a subjective point of view - building inadequate in size or qualitative characteristics -, in the sense that the benefit also applies in the case of the availability of accommodation that is not concretely suitable, in terms of size and overall characteristics, to meet the housing needs of the interested party. `` and in the same sense also Cass., Sec. 6-5, Order No. 5051 of 24/02/2021, Cass., Sec. 6-5, Order No. 18091 of 05/07/2019, and Cass., Sec. 6-5, Order No. 18092 of 05/07/2019). It is also worth recalling that the Constitutional Court itself, in declaring inadmissible the question of constitutional legitimacy of letter b) of number 1) of note II-bis of art. 1 of the first part of the tariff attached to the d.p.r. of April 26, 1986, no. 131 (raised in relation to articles 3 and 53 of the Constitution and to the principles of reasonableness, rationality and non-contradiction), with Order No. 203 of July 6, 2011, has expressly recognized that ``the concessions in question respond to the reasonable rationale of favoring the purchase of a dwelling in the place of residence or work to the advantage of those who, in the same place, do not have the possession of another dwelling house objectively suitable to meet their needs". \\
\midrule
\textbf{GPT3} & \small The acts of grievance are unfounded and must be rejected. The property pre-purchased by the applicant consists of a small room and a bathroom, with a total cadastral area of 25.00 square meters and a usable living area of 17.90 square meters. The technician has ascertained that the property shows traces of rising damp, mold, and fungi, concluding that it is not suitable for meeting the most basic housing needs for a family unit consisting of a father and two school-aged children. Law December 28, 1995, no. 549 establishes that for the recognition of the concession, in the purchase deed the buyer must declare not to be the exclusive holder or in communion with the spouse of the rights of property, usufruct, use and dwelling of another dwelling house in the territory of the municipality in which the property to be purchased is located. The prevailing jurisprudence of the supreme college has adhered to the interpretative option according to which the mere ownership of a real estate asset is not an obstacle to the recognition of the concession, which is instead granted to the taxpayer who does not own a property that can be used as a dwelling. \\
\midrule
\textbf{GPT4} & \small The fundamental preliminary question concerns the evaluation of the suitability or otherwise of the property pre-purchased by the appellant, with respect to the housing needs of himself and his family. It appears from the same that the property in question consists of a small room and a bathroom (depicted in the attached photographic documentation), having a total cadastral area of 25.00 sqm and a usable living area of 17.90 sqm (as shown in the attached plan). The technician also ascertained that the property shows traces of rising damp, mold and fungi, concluding that it is not suitable for meeting the most basic housing needs for a family unit composed of a father and two school-aged children. Based on the aforementioned assessment, supported by plans and photographs, it must therefore be considered established that the property purchased by the respondent\_1 on 19/3/2003, must be considered unsuitable for meeting the housing needs of the aforementioned and his family, so that the first judge correctly considered such purchase not preclusive with respect to the tax benefits claimed. Therefore, the appeal filed must be rejected and the contested judgment must be confirmed \\
\bottomrule
\end{longtable}
\end{center}

\subsection{Abstractive Summary of the Reasons for the Decision}
In this section, we present the abstractive summaries, distinguishing flowing-text ones and issue-based ones.

\subsubsection{Flowing-Text Summaries}\label{app:fabstr}

\begin{center}
\begin{longtable}{p{0.15\textwidth}|p{0.8\textwidth}}
\caption{Flowing-text summaries generated by LLMs.} \label{tab:flowing} \\

\hline \multicolumn{1}{c}{\textbf{Technique}} & \multicolumn{1}{c}{\textbf{Summary}} \\ \hline 
\endfirsthead

\multicolumn{2}{c}%
{{\textbf{\tablename\ \thetable{}} -- Flowing-text summaries generated by LLMs (Continued)}} \\
\hline \multicolumn{1}{c}{\textbf{Technique}} & \multicolumn{1}{c}{\textbf{Summary}}  \\ \hline 
\endhead

\multicolumn{2}{r}{{Continued on next page}} \\ \hline
\endfoot

\endlastfoot
\textbf{IT5} & \small In terms of tax relief for the first home, pursuant to art. 1 of the tariff attached to the d.p.r. 26.05.1992 n. 131 (ex art. 3, paragraph 131), the property purchased by the defendant\_1 is unsuitable to meet the housing needs of the same and his family, so the first judge correctly considered that the benefit applies even in the case of availability of a dwelling that is not concretely suitable, due to its small size, to be intended for his residence, to meet his housing needs, to satisfy his qualitative characteristics, for housing needs of the property (art. 549 cc. n.) for his housing needs (ex cp. 26.12.1997 cc. 548 cpc. 26.650 of the right to a property purchased by the taxpayer, so the benefit ex art. 141 of the taxpayer, as the buyer is inadequate, in case of a property that the dwelling that the enjoyment of the dwelling (and availability of housing) and cannot be intended to be used for their own housing needs) and for the real estate needs, for his family and is not suitable for his property. \\
\midrule
\textbf{GPT3} & \small The acts of grievance are unfounded and must be rejected. The fundamental preliminary issue is that concerning the assessment of the suitability or otherwise of the property pre-purchased by the applicant, with respect to the housing needs of himself and his family. According to the sworn expert report, the property consists of a small room and a bathroom, has a total cadastral area of 25 square meters and a usable living area of 17.90 square meters. The property shows traces of rising damp, mold and fungi, and is unsuitable for meeting the most basic housing needs for a family unit consisting of a father and two school-age children. Law No. 549 of December 28, 1995, establishes that in order to benefit from the tax relief, the buyer must declare that he is not the exclusive owner or in communion with the spouse of the rights of property, usufruct, use and habitation of another dwelling house in the territory of the municipality where the property to be purchased is located. The prevailing jurisprudence of the supreme college has adhered to the interpretative option according to which the mere ownership of a real estate asset is not an obstacle to the recognition of the relief, if the property is not concretely suitable to meet the housing needs of the interested party. Therefore, the appeal must be rejected and the contested judgment must be confirmed. The costs of this phase must be borne by the office. \\
\midrule
\textbf{GPT4} & \small The text concerns the issue of the suitability of a property pre-purchased by the applicant to meet the housing needs of his family. The sworn appraisal filed by the respondent highlights that the property, small in size and with traces of humidity, mold, and fungi, is not suitable for accommodating a family unit consisting of a father and two school-aged children. Therefore, the judge deemed it appropriate not to consider such a purchase as preclusive for the tax benefits claimed. The prevailing case law maintains that mere ownership of a property is not enough to deny the relief, which is instead granted to those who do not own a suitable dwelling. The appeal submitted is rejected and the contested judgment is confirmed, with the costs of this phase charged to the office. \\
\bottomrule
\end{longtable}
\end{center}

\subsubsection{Issue-Based}

\begin{center}
\begin{longtable}{p{0.15\textwidth}|p{0.8\textwidth}}
\caption{Issue-based summaries} \label{tab:issue} \\

\hline \multicolumn{1}{c}{\textbf{Technique}} & \multicolumn{1}{c}{\textbf{Summary}} \\ \hline 
\endfirsthead

\multicolumn{2}{c}%
{{\textbf{\tablename\ \thetable{}} -- Issue-based summaries (Continued)}} \\
\hline \multicolumn{1}{c}{\textbf{Technique}} & \multicolumn{1}{c}{\textbf{Summary}}  \\ \hline 
\endhead

\multicolumn{2}{r}{{Continued on next page}} \\ \hline
\endfoot

\endlastfoot
\textbf{GPT3} & \small
\textbf{QD1}: What is the fundamental preliminary question?

\textbf{PD1}: The fundamental preliminary question concerns the assessment of the suitability or otherwise of the property pre-purchased by the applicant, with respect to the housing needs of the applicant and their family.

\textbf{BT1}: ["Article 16 of the d.l. May 22, 1993, n. 155", ``the purchase deed the buyer declares not to own another building <<suitable for housing>> in the same municipality", ``the concept of ``suitability" of the pre-owned house - obstacle to the enjoyment of the benefit (and expressly provided for in the previous legislation) - must be considered intrinsic to the notion of ``dwelling house" itself"]

\textbf{QD2}: What is the rationale behind first-home benefits?

\textbf{PD2}: The rationale behind first-home benefits is to promote the purchase of a dwelling in the place of residence or work for the benefit of those who, in the same place, do not have possession of another objectively suitable dwelling house to meet their needs.

\textbf{BT2}: ["Article 3, paragraph 131, of the l. n. 549 of 1995", ``the concept of ``suitability" of the pre-owned house must be assessed both from an objective point of view - actual uninhabitability - and from a subjective point of view - building inadequate in size or qualitative characteristics -", ``the fact that the buyer of the property is also the owner of another property that is unsuitable, due to its small size, to be used as their dwelling is not an obstacle to the application of the first-home benefits"]

\textbf{KW}: ["first-home benefits", ``pre-purchased property", ``housing suitability", ``assessment", ``tax benefits"]\\
\midrule
\textbf{GPT4} & \small \textbf{QD1}: What is the criterion for determining whether a pre-purchased property is suitable or not to meet the housing needs of the owner and his family? 

\textbf{PD1}: The mere ownership of a real estate asset does not prevent the recognition of the concession, which is instead granted to the taxpayer who does not own a property that can be used as a dwelling, considering both the actual uninhabitability (objective profile) and the inadequacy in terms of size or qualitative characteristics (subjective profile) (Cass. civ n 20981/2021, Cass. sect. 5, order n. 19989 of 27/07/2018, Cass. sect. 5, judgment n. 2565 of 02/02/2018). 

\textbf{BT1}: [the property in question consists of a small room and a bathroom, with a total cadastral area of 25.00 sqm and a useful living area of 17.90 sqm] BT2: [the property shows traces of rising damp, mold, and fungi] BT3: [Article 16 of the d.l. 22 May 1993, n. 155, amending the t.u. registration tax, had prescribed, for the recognition of the concession, that the buyer declare in the purchase deed not to own another building ``suitable for housing" in the same municipality]

\textbf{QD2}: What is the current interpretation of the legislation on tax concessions for the first home in relation to the suitability of the pre-owned dwelling? 

\textbf{PD2}: The suitability of the pre-owned dwelling must be assessed both from an objective point of view (actual uninhabitability) and from a subjective one (inadequate building in terms of size or qualitative characteristics), meaning that the benefit also applies in the case of the availability of a dwelling that is not concretely suitable, in terms of size and overall characteristics, to meet the housing needs of the interested party (Cass., sect. 5, order n. 19989 of 27/07/2018, Cass., sect. 5, judgment n. 2565 of 02/02/2018). 

\textbf{BT2.1}: [on the subject of tax concessions for the first home, pursuant to art. 1, note ii bis, of the tariff attached to the d.p.r. n. 131 of 1986, in the text (applicable ``ratione temporis") amended by art. 3, paragraph 131, of the law n. 549 of 1995] 

\textbf{BT2.2}: [the concept of ``suitability" of the pre-owned home - an obstacle to the enjoyment of the benefit (and expressly provided for in the previous legislation) - must be considered intrinsic to the very notion of ``house of residence", to be understood as a dwelling concretely suitable, both from an objective-material and legal point of view, to meet the housing needs of the interested party] 

\textbf{BT2.3}: [the concessions under examination respond to the reasonable rationale of favoring the purchase of a dwelling in the place of residence or work for the benefit of those who do not have possession of another house of residence objectively suitable to meet their needs] 

\textbf{KW}: [tax concessions, first home, housing suitability, housing needs, uninhabitability, inadequacy, property ownership, legislation, jurisprudence] \\
\bottomrule
\end{longtable}
\end{center}